\documentclass{article}

\let\Pr\relax

\newcommand{\E}{\mathbb{E}}
\newcommand{\Pr}{\mathbb{P}}
\newcommand{\R}{\mathbb{R}}

\newcommand{\N}{\mathbb{N}}

\newcommand{\ind}{\mathbbm{1}}

\newcommand{\norm}[1]{\left\lVert #1 \right\rVert}

\newcommand{\argmax}{\text{arg\,max}}
\newcommand{\argmin}{\text{arg\,min}}
\newcommand{\ceil}[1]{\lceil {#1} \rceil}

\newcommand{\abs}[1]{\left| #1 \right|}

\newcommand{\dist}{\text{Dist}}

\usepackage[preprint]{neurips_2026}

\usepackage[utf8]{inputenc} 
\usepackage[T1]{fontenc}    
\usepackage{hyperref}       
\usepackage{url}            
\usepackage{booktabs}       
\usepackage{amsfonts}       
\usepackage{nicefrac}       
\usepackage{microtype}      
\usepackage{xcolor}         
\usepackage{hyperref}       
\hypersetup{
	colorlinks   = true, 
	urlcolor     = blue, 
	linkcolor    = blue, 
	citecolor   = blue 
}
\title{Multi-Objective Multi-Agent Bandits: From Learning Efficiency to Fairness Optimization}

\author{%
  John Wang\\
  University of Massachusetts Amherst \\
  \texttt{jawang@umass.edu} \\
  \And
  Mengfan Xu\\
  University of Massachusetts Amherst \\
  \texttt{mengfanxu@umass.edu} \\
}

\usepackage{amsmath,amsfonts,amsthm,amssymb}
\usepackage{natbib}
\usepackage{bbm}
\usepackage[ruled,vlined,linesnumbered,noend]{algorithm2e}
\newtheorem{theorem}{Theorem}
\usepackage{wrapfig}
\usepackage{graphicx}
\usepackage{nicefrac}
\begin{document}

\maketitle

\begin{abstract}
We study multi-objective multi-agent multi-armed bandits (MO-MA-MAB) under
stochastic rewards, where agents observe heterogeneous reward vectors and
communicate over time-varying graphs. We formulate this emerging problem setting to address \emph{efficient learning}, measured by Pareto regret, and incorporate \emph{fair learning} as an additional goal, captured via social welfare.  To measure
efficiency, we formulate Pareto regret and develop \textsc{Pareto UCB1 Gossip},
whose novel exploration radius explicitly separates statistical uncertainty in
Pareto-based inference from consensus error. To express the fairness constraint, we formulate a Nash
Social Welfare objective over preference-scalarized rewards and propose
\textsc{Simulated NSW UCB Gossip}, which integrates preference-based reward
simulation, gossip-based utility estimation, and UCB-style exploration. We prove
that \textsc{Pareto UCB1 Gossip} achieves \(\mathcal{O}(\log T)\) regret and an
instance-independent rate of \(\mathcal{O}(\sqrt{T})\), while \textsc{Simulated
NSW UCB Gossip} achieves an instance-independent regret bound of
\(\mathcal{O}(T^{3/4})\). This separation reveals the cost of imposing the fairness constraint to our efficiency objective: fairness limits information
aggregation and slows convergence. Experiments show that our methods
consistently outperform baselines, improving performance by approximately
\(100\%\) and \(50\%\) in the efficiency and fairness settings, respectively.\end{abstract}

\section{Introduction}

Sequential decision-making under uncertainty is commonly modeled using the multi-armed bandit (MAB) framework. In its classical form, a learner repeatedly selects from a set of actions (or “arms”), each associated with an unknown reward distribution, and observes only the reward of the chosen action. The learner’s objective is to maximize cumulative reward over time, despite two fundamental challenges: uncertainty in the underlying reward distributions and limited feedback, as observations are restricted to the selected actions.  These challenges give rise to the fundamental trade-off between exploration, gathering information about uncertain arms, and exploitation, leveraging current estimates to select the best-performing option.

This combination of uncertainty and partial observability arises naturally in many real-world applications. In online recommendation systems a platform must select a single item to display while only observing user feedback for that choice, leaving the outcomes of alternative recommendations unknown. In adaptive clinical trials, treatments must be sequentially assigned to patients with uncertain efficacy, and only the outcomes of administered treatments are observed. In networked systems, decisions such as resource allocation or routing must be made under stochastic conditions with feedback limited to realized actions. Across these domains, the need to learn from noisy, incomplete observations while optimizing performance makes the MAB framework a powerful and widely applicable modeling paradigm.

Many modern decision-making systems are inherently multi-agent, where multiple learners interact with a shared set of actions under uncertainty. This motivates the study of multi-agent multi-armed bandits (MA-MAB) in the stochastic setting, where each agent observes noisy rewards and must learn the underlying reward structure over time. In contrast to the single-agent case, agents can leverage information generated by others to accelerate learning, partially mitigating the challenges of uncertainty and limited feedback. By combining local observations with information shared across agents, MA-MAB frameworks enable the formation of more accurate and robust reward estimates, improving learning efficiency in the presence of noise. Such cooperative learning arises naturally in applications including distributed sensor networks aggregating noisy measurements, wireless communication systems where devices learn channel quality through shared experience, and large-scale recommendation platforms that pool feedback across users.

While cooperation across agents can significantly improve learning efficiency, it also introduces new challenges when agents experience heterogeneous rewards. In many settings, different agents may observe systematically different reward distributions for the same action due to variations in preferences, local environments, or contextual factors. This heterogeneity constitutes an additional source of uncertainty, beyond the inherent stochasticity of rewards. Unlike the homogeneous setting, where information can be directly aggregated, heterogeneity requires agents to reconcile potentially conflicting observations when forming estimates.

Another key dimension of multi-agent systems is the distinction between centralized and decentralized learning. In centralized settings, a global coordinator can aggregate observations from all agents, enabling full information sharing. In contrast, decentralized systems lack such coordination, and agents must rely on local communication, introducing an additional layer of limited information. This challenge is further amplified when communication is time-varying. While some works assume stationary networks with fixed connections, many real-world systems exhibit nonstationary communication, where links between agents evolve over time. A common approach to modeling such dynamics is through Erdos–Renyi (ER) random graphs, in which edges are formed probabilistically at each time step.

Building on these challenges, an emerging direction of research considers multi-objective multi-armed bandits (MO-MAB), where each action yields a vector-valued reward rather than a single scalar outcome. This formulation captures settings in which performance is inherently multi-dimensional, and decisions must account for multiple criteria simultaneously. By representing rewards as vectors, MO-MAB provides a more realistic framework for modeling real-world scenarios, where outcomes cannot be adequately summarized by a single metric.

Building on the multi-objective formulation, a more realistic setting arises when different agents assign different utilities to the components of the reward vector. In such cases, each agent evaluates outcomes according to its own preferences, leading to potentially conflicting objectives across the system. This introduces the need to balance these preferences in a principled and fair manner, rather than simply determining a globally optimal arm or optimizing cumulative regret. Instead, it becomes natural to consider social welfare functions that aggregate individual utilities, providing a unified measure of system performance that accounts for both efficiency and fairness.

These challenges highlight the need for new learning frameworks that account
for both multi-objective structure and fairness constraints, motivating the
following research questions: (1) Can the multi-agent multi-armed bandit
(MA-MAB) framework be extended to the multi-objective setting, incorporating
heterogeneous rewards and time-varying decentralized communication, while still
enabling efficient learning? (2) How can we formalize fairness in the
MO-MA-MAB setting through agent-specific utilities and design algorithms that
optimize a principled social welfare objective, such as Nash Social Welfare?
We answer these questions affirmatively through the following contributions.

\subsection{Contributions}


To address efficient learning in MO-MA-MAB, we develop a decentralized algorithm by extending the \textsc{Pareto UCB1} framework introduced in \citep{drugan2013designing} to a setting with heterogeneous rewards and time-varying decentralized communication. We incorporate a gossip-based communication protocol over Erdos–Renyi networks, as studied in \citep{liu2025distributed} to enable information sharing across agents. A key technical contribution is the design of an exploration radius that captures both sampling error and consensus error arising from decentralized estimation under heterogeneity, enabling the construction of reliable confidence sets for learning.

To enable fairness-aware efficent learning in MO-MA-MAB, we extend the notion of Nash Social Welfare (NSW) regret introduced in \cite{hossain2021fair} to the multi-objective multi-agent setting. We model agent-specific preferences through preference vectors that scalarize vector-valued rewards, so that each entry of the utility matrix—representing the mean reward of an arm for a given agent—is expressed as a linear combination of reward components weighted by that agent’s preferences. Building on this formulation, we develop a decentralized algorithm by extending the UCB-based approach in \citep{hossain2021fair} and incorporating gossip-based updates over Erdos–Renyi networks as in \citep{liu2025distributed}. Our method enables agents to optimize NSW with respect to the global average utility by leveraging consensus information, despite decentralized communication and heterogeneous observations.

We provide theoretical guarantees for both proposed algorithms through regret analysis. For \textsc{Pareto UCB1 Gossip}, we establish a logarithmic regret bound of order $\mathcal{O} (\log T)$, which implies an instance-independent bound of $\mathcal{O} (\sqrt{T} )$, matching the optimal rate for stochastic bandits. In contrast, for \textsc{Simulated NSW UCB Gossip}, we derive a regret bound of order $\mathcal{O} (T^{3/4})$, reflecting the additional complexity introduced by optimizing the nonlinear Nash Social Welfare objective under heterogeneous rewards. This gap highlights a fundamental tradeoff between efficiency and fairness, as enforcing fairness prevents the aggregation of observations across agents and slows learning.

We complement our theoretical results with empirical evaluations against natural baselines. In the multi-objective setting, we compare against a multi-agent extension of \textsc{Pareto UCB1} introduced in \citep{drugan2013designing} and a multi-objective extension of \textsc{Gossip Successive Elimination} developed in \citep{liu2025distributed}, demonstrating faster convergence and improved regret due to more efficient confidence bounds. In the social choice setting, we compare against variants that remove either gossip communication or reward simulation, showing that both components are essential for achieving sublinear regret. We review related work in Appendix \ref{app:related_work}. 

\section{Problem Formulation}

The multi-objective, multi-agent, multi-armed bandit problem features $N$ agents, $K$ arms, $D$ dimensional rewards, and a time horizon $T$. At time $t\in [T]$, each agent $i\in[N]$ pulls an arm $a_i^t\in [K]$. The rewards associated with agent $i$ arm $k$, $X_{i,k}(t)\in [0,1]^D$ are i.i.d across time $t\in [T]$ and are sub-Gaussian distributed with mean $\mu_{i,k}$. We assume a heterogeneous setting where $\mu_{i,k}\neq\mu_{j,k}$ for $i\neq j$. We introduce $\mu_k =\frac{1}{N}\sum_{i=1}^N\mu_{i,k}$ as the centered mean for arm $k$.

Our problem follows an decentralized communication model. Specifically, we introduce an undirected base graph $\mathcal{G} =(V,E)$ where $V=[N]$, the set of agents and for $i,j\in [N]$, $(i,j)\in E$ if and only if agent $i$ can communicate with agent $j$. $\mathcal{G}$ is assumed to be connected and may in certain instances be complete. A random graph $\mathcal{G}_t$ restricts communication between agents at time step $t\in [T]$. $\mathcal{G}_t$ is generated according to the Erdos-Renyi model by selecting edges from the base graph $\mathcal{G}$ with probability $p$, a hyperparameter. At time $t$, the graph $\mathcal{G}_t$ corresponds to a weight matrix $W_t\in\R^{N\times N}$. We define $W_t$ by its entries
$[W_t]_{ij}=\begin{cases}
    \nicefrac{1}{N} & \text{if } j\in\mathcal{N}_i(t) \\
    1-\nicefrac{\abs{\mathcal{N}_i(t)}}{N} & \text{otherwise}
\end{cases}$.
In other words, agent $i$ assigns each of its neighbors $\frac{1}{N}$ of the weight and applies the rest of the mass to itself. Note that $W_t$ is row stochastic with spectral gap parameter $\rho =\sum_t\E [\norm{W_t -\frac{1}{N}\ind\ind^\top}_2 ]<1$.

\subsection{Multi-Objective Pareto Setting}

We extend cumulative expected regret depicted in \citep{liu2025distributed} to MO-MA-MAB using Pareto optimality. As discussed in \citep{drugan2013designing}, for vectors $x,y\in\R^D$, $x$ dominates $y$ and we write $x\succ y$ if for all $i\in [D]$, $x_i\geq y_i$ and there exists an $i\in [D]$ such that $x_i>y_i$. The $\epsilon$-distance between a vector $x\in\R^D$ and a set $S$ of vectors with the sample dimensionality is given by
$\text{Dist} (x,S)=\inf\{\epsilon >0:\nexists y\in S\text{ such that } y\succ x+\epsilon\}$. 
In other words, $\epsilon$ is the smallest relaxation needed so that no vector $y\in S$ dominates $x$. Similarly, a vector $y\in S$ is Pareto optimal if it is not dominated by another vector in $S$. We express global Pareto regret for MO-MA-MAB as
$R_T=\sum_{i=1}^{N}\sum_{t=1}^{T} \dist (\mu_{a_i^t} ,O)$ 
where $\mu_{a_i^t}$ is the centered arm mean for the arm selected by agent $i$ at time $t$ and $O=\{\mu_{k^*} :\nexists k\in [K]\text{ such that } \mu_k\succ\mu_{k^*}\}$. We refer to this set as the Pareto front of the centered arm means because it contains all Pareto optimal vectors. In other words, the sub-optimality gap for global Pareto regret is the $\epsilon$-distance between the centered mean of the arm selected and the Pareto front.

\subsection{Nash Social Welfare Setting}

A constant preference vector $w_i\in\Delta_K$ denotes the utility of agent $i$ where $\Delta_k$ refers to the probability simplex over the arm set. When agent $i$ pulls arm $k$ at time $t$, it receives the scalar reward $w_i^\top X_{i,k}(t)$, the dot product between the vector reward and its utility. We represent the scalar mean for agent $i$ pulling arm $k$ with agent $j$'s preferences, $\mu_{i,j,k}^*=w_j^\top\mu_{i,k}$. The centered scalar mean for arm $k$ with agent $j$'s preferences is $\bar{\mu}_{j,k}=\frac{1}{N}\sum_{i=1}^N\mu_{i,j,k}^*$. 

Our regret formulation follows from the Nash Social Welfare function to develop fair algorithms for MO-MA-MAB. Given a distribution over the arm set $p\in\Delta_K$ and a utility matrix $\mu\in\R^{N\times K}$, Nash Social Welfare is given by
$\text{NSW}(p,\mu )=\prod_{j=1}^N\left(\sum_{k=1}^K p_k\cdot \mu_{j,k}\right)$
Suppose that the optimal distribution under the centered scalar mean matrix is $p^*=\argmax_{p\in\Delta_K}\text{NSW} (p,\bar{\mu} )$. Then we define NSW regret as
$R_T=\sum_{t=1}^T\sum_{i=1}^N\text{NSW} (p^*,\bar{\mu} )-\text{NSW} (p_i^t,\bar{\mu} )$
The suboptimality gap for agent $i$ at time $t$ is the difference between the NSW of the optimal distribution $p^*$ under the centered scalar mean matrix $\bar{\mu}$ and the estimated distribution $p_i^t$ under the centered scalar mean matrix. NSW regret is calculated by summing the suboptimality gaps over agents $i$ and time $t$. The decision to adopt centered scalar means as the reference utility matrix parallels \citep{liu2025distributed}. We interpret reward heterogeneity as stochastic noise retaining the representation of conflicting utilities with different preference vectors. Since communication is restricted, our regret formulation allows agents to settle at different estimated distributions at each time step.

Our problem setting extends similar works that study MA-MAB using NSW regret. \citep{hossain2021fair} and \citep{jones2023efficient}, considers a utility matrix specifying the rewards for each agent under a homogeneous, centralized setting. \citep{zhang2024no} generalizes the problem to non-stationary utility matrices. Our work extends the problem by incorporating multi-dimensional rewards. Specifically, we build on \citep{hossain2021fair} to the case where each element in the utility matrix is a linear combination of the dimensions of the reward weighted by a preference vector. Our work also broadens \cite{hossain2021fair} to the setting where rewards are heterogeneous and communication is decentralized.

\section{Pareto UCB1 Gossip}

To enable efficient learning in the multi-objective Pareto setting, we propose an algorithm that extends \textsc{Pareto UCB1} \citep{drugan2013designing} to the multi-agent regime by integrating gossip-based sharing, as in \textsc{Successive Gossip Elimination} \citep{liu2025distributed}. To address the challenges introduced by heterogeneous reward vectors and decentralized communication, we develop an exploration term that captures both sampling and consensus errors. 

\subsection{Methodology}

At each time step, every agent maintains local averages of arm rewards together with a set of global averages that are updated through gossip-based communication. The algorithm begins with an initialization phase in which each arm is pulled once to obtain initial samples. After this, each agent constructs upper confidence bounds for all arms using its global averages and a confidence radius which accounts for both sampling uncertainty and consensus error. Using the upper confidence bounds, each agent identifies the Pareto front and selects an arm uniformly at random from this set. Upon receiving a reward, the agent updates its local empirical mean for the selected arm and increments the corresponding counter. Finally, agents perform a gossip update, mixing their current estimates with those of their neighbors according to the time-varying communication matrix, while incorporating the latest local updates. This iterative process enables agents to progressively refine their estimates by combining local observations with information propagated through the network, leading to improved identification of Pareto optimal arms over time. A complete description is presented in Algorithm \ref{alg:dr}. 

\begin{algorithm}[ht]
    \SetAlgoLined
    \caption{Pareto UCB1 Gossip}
    \label{alg:dr}
    
    \textbf{Input:} total time horizon \(T\), set of agents \([N]\), set of arms \([K]\), dimensions \([D]\), base graph \(\mathcal{G}\), link probability \(p\)\;
    
    \textbf{Initialization:} For each agent \(i\in [N]\) and arm \(k\in [K]\), set \(T_{i,k}(0)=0\), \(\hat{\mu}_{i,k}(0)=0\), and \(z_{i,k}(1)=0\)\;
    
    \For{\(t=1,\dots,T\)}{
        \For{\(i=1,\dots,N\)}{
            \If{\(t=1,\dots,K\)}{
                \(a_i^t\gets t\)\;
            }
            \Else{
                \(\text{UCB}_{i,k}(t)\gets z_{i,k}(t)+c_{i,k}(t)\), where 
                \(c_{i,k}(t)=\sqrt{\nicefrac{8\log \left(t\sqrt[4]{D\left|\mathcal{A}^*\right|}\right)}{T_{i,k}(t)}}+\nicefrac{2\sqrt{n}}{1-\sqrt{p}}\)\;
                
                \(S_i^t=\{k^*\in [K]: \nexists k\in [K]\text{ such that } \text{UCB}_{i,k}(t)\succ \text{UCB}_{i,k^*}(t)\}\)\;
                
                Sample \(a_i^t\) uniformly from \(S_i^t\)\;
            }
            
            Observe \(X_{i,a_i^t}(t)\sim\mu_{i,a_i^t}\); 
            \(T_{i,a_i^t}(t)\gets T_{i,a_i^t}(t-1)+1\)\;
            
            \(\hat{\mu}_{i,a_i^t}(t)\gets
            \nicefrac{T_{i,a_i^t}(t-1)\hat{\mu}_{i,a_i^t}(t-1)+X_{i,a_i^t}(t)}
            {T_{i,a_i^t}(t)}\)\;
            
            \For{\(k=1,\dots,K\)}{
                \(z_{i,k}(t+1)\gets
                \sum_{j=1}^N[W_t]_{ij}z_{j,k}(t)
                +\hat{\mu}_{i,k}(t)-\hat{\mu}_{i,k}(t-1)\)\;
            }
        }
    }
\end{algorithm}
\subsection{Analysis}

We analyze the regret of \textsc{Pareto UCB1 Gossip} by extending the \textsc{Pareto UCB1} framework to a decentralized, heterogeneous setting. The main challenge is to control how statistical uncertainty and imperfect information mixing jointly affect the confidence bounds used for arm selection. The comprehensive proof is provided in Appendix \ref{proof:mo}. 

\subsubsection{Key Decomposition and Error Bounds}

We have $z_{i,k}(t)$ as the global average of agent $i$ for arm $k$. We decompose the deviation between the global average and the centered arm mean as
$z_{i,k}(t)-\mu_k =(z_{i,k}(t) -\tilde{\mu}_k(t) )+(\tilde{\mu}_k (t) -\mu_k ),$
where $\tilde{\mu}_k (t)=\frac{1}{N}\sum_{i=1}^N\hat{\mu}_{i,k}$ is the average of local empirical means. The second term (sampling error) concentrates via Chernoff bounds, yielding exponential decay in $s_k(t)=\min_i T_{i,k}(t)$. The first term (consensus error) is controlled through the contraction of the gossip updates, leading to a uniform bound of order $\frac{\sqrt{N}}{1-\sqrt{\rho}}$ with high probability. Combining both bounds gives
$\Pr [z_{i,k}(t)\nprec\mu_k +\epsilon\mathbf{1} ]\leq D\exp\left(- \tfrac{1}{2} s_k(t)\epsilon^2\right) +\frac{N^2}{t}.$

A key novelty is the resulting exploration radius
$$c_{i,k}(t)=\sqrt{\nicefrac{8\log (t\sqrt[4]{D\abs{\mathcal{A}^*}} )}{T_{i,k}(t)}} +\nicefrac{2\sqrt{N}}{1-\sqrt{p}},$$
which cleanly separates sampling and consensus effects. In contrast to prior gossip-based approaches where these terms are coupled, the sampling component retains the optimal $\mathcal{O}(\sqrt{\log T/T_{i,k}(t)})$ decay, while the consensus error appears only as an additive constant. This distinction is critical for ensuring that confidence intervals shrink at the correct rate.

\subsubsection{Regret Bound}

We now outline how the regret bound is obtained. The main objective is to bound the expected number of times a suboptimal arm $k\notin\mathcal{A}^*$ is selected by an agent $i$. After a warm-up period, a suboptimal arm can only be selected if its upper confidence bound is not dominated by that of some Pareto optimal arm $h\in\mathcal{A}^*$. This implies that at least one of the following failure events occurs:
$$\text{(i)}\; z_{i,h}(t)\nsucc\mu_h -c_{i,h}(t)\mathbf{1}\quad ;\quad\text{(ii)}\; z_{i,k}(t)\nprec\mu_k +c_{i,k}(t)\mathbf{1}\quad ;\quad\text{(iii)}\;\mu_h\nsucc\mu_k +2c_{i,k}(t)\mathbf{1}.$$

The exploration radius governs these events in two ways. First, it determines the warm-up period: once $T_{i,k}(t)$ is sufficiently large so that $2c_{i,k}(t)\leq\Delta_k$, event (iii) cannot occur. Second, events (i) and (ii) are precisely the deviation events controlled in the previous subsection, and thus occur with probability at most $\mathcal{O} (t^{-4})+\mathcal{O} (N^2/t)$.

Combining these bounds yields a logarithmic upper bound on the expected number of pulls of each suboptimal arm. Summing over suboptimal arms and agents gives the global Pareto regret.

\begin{theorem}[Global Pareto Regret of \textsc{Pareto UCB1 Gossip}]
Let $\mathcal{A}^*$ denote the Pareto optimal set and $\Delta_k$ the Pareto gap of arm $k$. Then
$\mathrm{Reg}_T (\pi )\leq N\sum_{k\notin\mathcal{A}^*}\frac{32\log (T\sqrt[4]{D\abs{\mathcal{A}^*}} )}{\Delta_k} +N\left( 1+\frac{\pi^4}{45} +2\abs{\mathcal{A}^*} N^2(\log T+1)\right)\sum_{k\in\mathcal{A}^*}\Delta_k .$
\end{theorem}

Since the regret admits a logarithmic, gap-dependent bound of order $\mathcal{O} (\log T)$, standard arguments imply a corresponding gap-independent bound of order $\mathcal{O} (\sqrt{T} )$.

\subsubsection{Discussion and Comparison}

The leading term scales as $\mathcal{O} (\log T)$ with standard gap dependence, matching \textsc{Pareto UCB1}. The additional $\mathcal{O} (N^3\log T)$ contribution reflects the cost of decentralized communication through consensus error and diminishes with improved connectivity.

Compared to \citep{drugan2013designing}, we retain the centralized regret structure while extending it to decentralized, heterogeneous multi-agent settings. Compared to gossip-based bandits such as \citep{liu2025distributed}, the key improvement lies in the exploration radius: by separating sampling and consensus errors, it preserves the optimal $\mathcal{O} (\sqrt{\log T/T_{i,k}(t)})$ shrinkage while incorporating consensus effects only additively. This leads to faster elimination of suboptimal arms, with decentralization affecting only lower-order terms, consistent with empirical observations. 

\section{Simulated NSW UCB Gossip}

To optimize learning under fairness constraints in the Nash Social Welfare setting, we develop an algorithm that builds on the UCB-based approach in \citep{hossain2021fair} and incorporates gossip-based updates as in \textsc{Successive Gossip Elimination}. Our method leverages shared preference vectors among neighboring agents by constructing simulated rewards from the perspective of other agents.

\subsection{Methodology}

\begin{algorithm}[h]
\SetAlgoLined
\SetAlgoNoEnd
\caption{Simulated NSW UCB Gossip}
\label{alg:dr_NSW}

\textbf{Input:} total time horizon \(T\), set of agents \([N]\), set of arms \([K]\), dimensions \([D]\), preference vectors \(w_i\), base graph \(\mathcal{G}\), link probability \(p\), explore factor \(\alpha^t\)\;

\textbf{Initialization:} For each agent pair \(i,j\in [N]\) and arm \(k\in [K]\), set \(T_{i,j,k}(0)=0\), \(\hat{\mu}_{i,j,k}(0)=0\), and \(z_{i,j,k}(1)=0\)\;

\For{\(t=1,\dots,T\)}{
    \For{\(i=1,\dots,N\)}{
        \If{\(t=1,\dots,K\)}{
            \(p_i^t \gets e_t\)\;
        }
        \Else{
            \(p_i^t \gets \argmax_{p\in\Delta_K}\text{NSW}(p,z_i(t))
            +\alpha^t\sum_{k=1}^K p_k r_k^t\), where
            \(r_k^t=\sqrt{\nicefrac{\log(NKt)}{T_{i,j,k}(t)}}\)\;
        }

        Sample \(a_i^t\sim p_i^t\) and observe \(X_{i,a_i^t}(t)\)\;

        \For{\(j\in\mathcal{N}_i(t)\)}{
            \(s_{i,j}(t)\gets w_j^\top X_{i,a_i^t}(t)\);
            \(T_{i,j,a_i^t}(t)\gets T_{i,j,a_i^t}(t)+1\)\;

            \(\hat{\mu}_{i,j,a_i^t}(t)
            \gets
            \nicefrac{
            T_{i,j,a_i^t}(t-1)\hat{\mu}_{i,j,a_i^t}(t-1)+s_{i,j}(t)
            }{
            T_{i,j,a_i^t}(t)
            }\)\;
        }

        \For{\(j=1,\dots,N\)}{
            \For{\(k=1,\dots,K\)}{
                \(z_{i,j,k}(t+1)
                \gets
                \sum_{l=1}^N [W_t]_{il}z_{l,j,k}(t)
                +\hat{\mu}_{i,j,k}(t)-\hat{\mu}_{i,j,k}(t)\)\;
            }
        }
    }
}
\end{algorithm}
Algorithm \ref{alg:dr_NSW} can be understood per agent $i$ at a specific time $t$. Each round consists of three stages: choosing an arm, observing a rewards, and updating estimators. In the arm choosing period, during the first $K$ time steps the agent sets its distribution to the $t$th standard basis vector. This conforms to the exploration period in traditional UCB algorithms where every arm is pulled once. Beyond the first $K$ rounds, the agent optimizes the objective
$\argmax_{p\in\Delta_K}\text{NSW} (p,z_i(t))+\alpha^t\sum_{k=1}^K p_k\cdot r_k^t\text{ where } r_k^t=\sqrt{\nicefrac{\log (NKt)}{T_{i,j,k}(t)}}$
with respect to its global averages. The first term exploits the fairest distribution over the arm set using NSW. The second term represents an explore term weighted by a exploration factor $\alpha^t$. Each term $k$ in the sum incentivizes or decentivizes the agent to assign probability mass on $p_k$ given $r_k^t$ which is inversely proportional to the counter $T_{i,j,k}(t)$.

When observing a reward, the agent samples an arm from its estimated distribution. Then, the agent pulls on the arm and obtains a scalar reward given by the dot product of its preference vector and the vector reward. The agent also receives $\abs{\mathcal{N}_i (t)}$ simulated rewards arising from the dot product between each neighbor's preference vector and the observed reward. 

During the estimator update phase, agent $i$ stores both a local average matrix $\hat{\mu}_i\in\R^{N\times K}$ and a global average matrix $z_i\in\R^{N\times K}$ for the value of arm $k$ under agent $j$'s preferences. The local average represents a Monte-Carlo estimate over the locally observed scalar rewards. The global average is computed by taking weighted averages of the local averages. For the local average matrix, the agent updates the Monte-Carlo estimates for the arm that it selected under its own preferences and its neighbors preferences. For clarity, consider the local average as a $N\times N\times K$ tensor.
$$\setlength{\arraycolsep}{2pt}
\begin{bmatrix}
    \begin{bmatrix}
        \hat{\mu}_{1,1,1} (t) & \cdots & \hat{\mu}_{1,1,K} (t) \\
        \vdots & & \vdots \\
        \hat{\mu}_{1,N,1} (t) & \cdots & \hat{\mu}_{1,N,K} (t)
    \end{bmatrix} &
    \begin{bmatrix}
        \hat{\mu}_{2,1,1} (t) & \cdots & \hat{\mu}_{2,1,K} (t) \\
        \vdots & & \vdots \\
        \hat{\mu}_{2,N,1} (t) & \cdots & \hat{\mu}_{2,N,K} (t)
    \end{bmatrix} &
    \cdots &
    \begin{bmatrix}
        \hat{\mu}_{N,1,1} (t) & \cdots & \hat{\mu}_{N,1,K} (t) \\
        \vdots & & \vdots \\
        \hat{\mu}_{N,N,1} (t) & \cdots & \hat{\mu}_{N,N,K} (t)
    \end{bmatrix}
\end{bmatrix}$$
In this way, the agent can only update select entries in the row corresponding to the arm it selected that round. In the global estimate for arm $k$ under agent $j$'s preferences, agent $i$ averages the global averages of all other agents weighted by the the $i$th row of $W_t$. Finally, the agent adds in any local update contributions. Again, it is helpful to view the global average as a $N\times N\times K$ tensor.
The resulting global average matrix for agent $i$ is a linear combination of the sub-matrices in the tensor weighted by the $i$th row of $W^t$. In he end, we add local average matrix update at time $t$.

\subsection{Analysis}

We analyze the cumulative Nash Social Welfare (NSW) regret of \textsc{Simulated NSW UCB Gossip}. Our analysis decomposes the sources of error arising from stochastic rewards and decentralized communication, and shows how these propagate through the nonlinear NSW objective. The comprehensive proof is provided in Appendix \ref{app:proof_NSW}. 

\subsubsection{Key Decomposition and Error Bounds}

The proof begins by decomposing the deviation between the global estimates and the centered mean as
$z_{i,j,k}(t)-\bar{\mu}_{j,k} =(z_{i,j,k}(t)-\bar{z}_{j,k} (t))+(\bar{z}_{j,k} (t)-\bar{\mu}_{j,k} ),$ which separates the error into a consensus term induced by decentralized communication and a sampling term arising from stochastic rewards. A key observation is that the network average tracks the empirical mean,
$\bar{z}_{j,k} (t)=\frac{1}{N}\sum_{i=1}^N\hat{\mu}_{i,j,k} (t-1),$
so that $\bar{z}_{j,k} (t)$ is an unbiased estimator of $\bar{\mu}_{j,k}$. The sampling error is controlled via concentration of the local estimators together with a coupling between off-diagonal and diagonal counts under the Erdos–Renyi communication model, yielding a bound of order $\frac{1}{\sqrt{p}}\sqrt{\frac{\log(NKt)}{T_{i,i,k}(t)}}$. The consensus error follows from the contraction of the gossip process and accumulates geometrically with rate $\rho$.

\subsubsection{Regret Bound}

To connect estimation error to regret, we exploit a telescoping argument that converts the multiplicative NSW objective into an additive bound:
$\abs{\text{NSW}(p,\bar{\mu} )-\text{NSW}(p,z_i(t))}\leq\sum_{j=1}^N\sum_{k=1}^K p_k\abs{z_{i,j,k}(t) - \bar{\mu}_{j,k}}.$
This yields a high-probability event under which the discrepancy between the true and estimated objectives is controlled by a UCB-style exploration term aligned with the algorithm, together with lower-order corrections. Under this event, we compare the optimal distribution $p^*$ with the chosen distribution $p_i^t$ and reduce regret to the accumulation of sampling and consensus errors over time.

\begin{theorem}[NSW Regret of \textsc{Simulated NSW UCB Gossip}]
With probability at least $1-\mathcal{O} ((N^2KT)^{-2})$, the cumulative NSW regret satisfies
$R_T\leq 
\mathcal{O}\left(\frac{N^2K}{\sqrt{p}} \sqrt{\log (NKT)}\cdot T^{3/4}\right) +\mathcal{O}\left( N^2K\log^{3/2} T\right) +\mathcal{O}\left(\frac{N^3}{1-\rho}\log T\right) .$
\end{theorem}

The leading term scales as $\mathcal{O} (N^2K T^{3/4}/\sqrt{p} )$, while the remaining terms are lower-order contributions due to limited exploration and decentralized communication.

\subsubsection{Discussion and Comparison}

The dominant contribution to regret is the sampling term, which scales as $\mathcal{O} (\frac{N^2K}{\sqrt{p}} T^{3/4})$. The $T^{3/4}$ rate reflects the difficulty of optimizing the nonlinear NSW objective, while the $\frac{1}{\sqrt{p}}$ dependence captures the reduction in effective sample size due to sparse communication. The remaining terms are lower order: the $\mathcal{O} (N^2K\log^{3/2} T)$ term corresponds to a burn-in phase, and the $\mathcal{O} (\frac{N^3}{1-\rho}\log T)$ term reflects the cost of decentralized consensus, improving with faster mixing.

The $\mathcal{O} (T^{3/4})$ dependence in our regret bound is driven by a fundamental sampling term
$\sum_{t=1}^T\sum_{k=1}^K \nicefrac{p_{i,k}^t}{\sqrt{T_{i,i,k}(t)}},$
which cannot be improved to yield a $\sqrt{T}$ rate under heterogeneous rewards. In contrast, UCB-based MA-MAB algorithms \citep{hossain2021fair, jones2023efficient} for NSW regret in homogeneous, centralized settings achieve $\mathcal{O} (\sqrt{T})$ regret by effectively pooling observations across agents, allowing counters to scale with the total number of samples. In our setting, however, because $\mu_{i,k}\neq \mu_{j,k}$, observations cannot be aggregated across agents, and each agent must rely on its own diagonal counters $T_{i,i,k}(t)$ to estimate arm values. This prevents the effective sample size from scaling with $N$ and slows the growth of $T_{i,i,k}(t)$ along frequently sampled arms. Consequently, even under optimal allocation, the cumulative contribution of the inverse-square-root term scales as $T^{3/4}$ up to logarithmic factors, making this rate unavoidable in our setting.

This behavior parallels non-stationary bandit problems, where past observations cannot be fully reused and similar $T^{3/4}$ regret rates arise. For instance, sliding-window and restart-based UCB methods achieve $\mathcal{O} (T^{3/4})$ regret in abruptly changing environments \citep{garivier2011upper}, and variation-budget approaches yield comparable rates when the amount of non-stationarity is large or unknown \citep{besbes2014stochastic}. In both cases, limited sample reuse constrains learning in much the same way as heterogeneity does here. Finally, while our problem-dependent analysis yields a $\mathcal{O} (T^{3/4})$ regret bound, this stands in contrast to the instance-free $\mathcal{O} (\sqrt{T} )$ guarantee achieved by the algorithm in the previous section; this gap reflects the inherent tradeoff required to enforce fairness through the Nash Social Welfare objective.

\section{Experiments}

The primary hyperparameters for the simulation are the number of agents $N$, arms $K$, and the dimensions $d$. The environment consists of preference vectors for each agent and distributions for each arm. Each preference vector $w_i$ is drawn from a Dirichlet distribution with concentration parameter $\mathbf{\alpha} =\mathbf{1}_N$. It follows that $w_i$ is drawn uniformly from the probability simplex over the arm set. We simulate each arm distribution as a Bernoulli trial. To generate arm means, we sample base means $\mu_{k,d}^B\sim\text{Uniform} (0.2,0.8)$ for every arm $k$ and dimension $d$. We add heterogeneity by perturbing these base means for each agent. For each agent $i$, $\mu_{i,k,d} =\mu_{k,d}^B +p_{i,k,d}$ where $p_{i,k,d}\sim\text{Uniform} (-\verb|het_scale| ,\verb|het_scale|)$. \verb|het_scale| represents an additional hyperparameter in the experiment. We clip $\mu_{i,k,d}$ to the range $[0.05,0.95]$ so that the arm means lie between $0$ and $1$. The base graph $\mathcal{G}$ is complete and we determine $\mathcal{G}_t$ using the ER model with parameter $p$.

\subsection{Pareto UCB1 Gossip}

We compare the performance of \textsc{Pareto UCB1 Gossip} against two baselines. The first baseline extends \textsc{Pareto UCB1} described in \citep{drugan2013designing} to the multi-agent case by taking $N$ independent instances. The second baseline extends \textsc{Gossip Successive Elimination} introduced in \citep{liu2025distributed} to the multi-objective setting. As in \textsc{Gossip Successive Elimination}, agent $i$ selects the arm in the active set with the smallest counter value. To account for vector rewards, our local and global averages are 3D tensors in $\R^{N\times K\times D}$. Local updates are made element-wise and global averages for agent $i$ arm $k$ are computed by taking linear combinations of vectors rewards for arm $k$ weighted by the $i$th row of $W_t$. In order to eliminate arms from the active set, we compute confidence intervals by calculating half-widths. We obtain upper and lower bounds by combining these half-widths to the global averages element-wise. Analogous to \textsc{Gossip Successive Elimination}, we eliminate arm $k$ if its upper bound is dominated by the Pareto front constructed by all lower bounds for agent $i$. 

\begin{wrapfigure}{r}{0.42\textwidth}
    \centering
    \vspace{-11pt}
    \includegraphics[width=\linewidth]{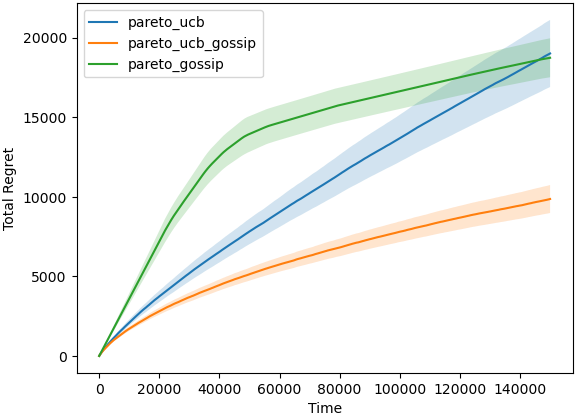}
    \vspace{-11pt}
\end{wrapfigure}

In the experiment, we set $N=8$, $K=8$, $D=3$, and $T=150000$. We set the graph connectivity $p=0.8$ and \verb|het_scale=0.2|. We perform $10$ trials, each corresponding to an independent environment with new arm means and connectivity graphs. For each trial, the algorithms run in the same environment but use different random number stream during the experiment when selecting an arm or sampling a reward. We graph the mean global Pareto regret along with confidence intervals set to one quarter of the standard deviation. In the legend, \verb|pareto_ucb_gossip| refers to \textsc{Pareto-UCB1 Gossip}, \verb|pareto_ucb| corresponds to baseline $1$, and \verb|pareto_gossip| refers to baseline $2$. Qualitatively, we observe that \verb|pareto_gossip| experiences a longer burn-in time due to confidence intervals converging slowly. On the other hand, we showed that the exploration term in \textsc{Pareto-UCB1 Gossip} converges on the order of $\mathcal{O} (\sqrt{\log T/T_{i,k}(t)})$. Finally, \verb|pareto_ucb| does not grow sub-linearly as the local averages are not unbiased estimators of the centered arm means.

\subsection{Simulated NSW UCB Gossip}

We analyze the performance of \textsc{Simulated NSW UCB Gossip} against two baselines. The first baseline \verb|no_gossip| follows the proposed algorithm but does not engage in gossip sharing. Instead, each agent uses its local average matrix to optimize NSW regret for arm selection. The local average matrix can still be computed as agents retain the ability to share preference vectors and conduct simulations using their own observations. The second baseline, referred to as \verb|no_sim|, mirrors the proposed algorithm without simulated rewards. Agents use the global average matrix to determine the arm distribution for each round. In the absence of reward simulation, agent $i$ can only compute the $i$th row of its local average matrix. Agents can populate their entire global average matrix via gossip. 

In this experiment, we consider $N=4$ agents, $K=5$ arms, feature dimension $D=2$, and horizon $T=1500$. The communication graph is generated with connectivity parameter $p=0.5$, and \verb|het_scale=0.2|. We run 15 independent trials, each corresponding to a newly sampled environment with fresh arm means. Within each trial, all algorithms share the same environment but use independent randomness for graph generation, arm selection, and reward sampling. We report the mean NSW regret with confidence intervals given by one quarter of the standard deviation.

\begin{center}
\vspace{-11pt}
    \includegraphics[scale=0.4]{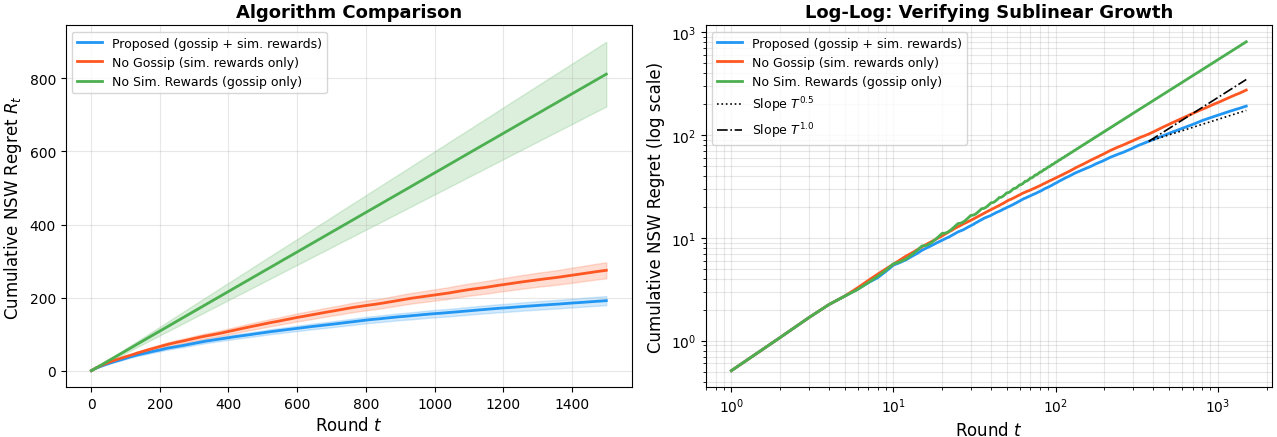}
    \vspace{-11pt}
\end{center}

We set the time horizon to $1500$ times steps as the proposed algorithm had already converged for most trials. Empirical results suggest that \verb|no_gossip| and \verb|no_sim| fail to exhibit sublinear growth. This observation is justified, as neither baseline produces an unbiased estimator of the centered means. In \verb|no_gossip|, each agent computes local averages using only samples from its own reward distributions. In \verb|no_sim|, agent $i$ receives global averages for agent $j$ computed with agent $j$'s reward distributions.

\bibliographystyle{abbrv}
\bibliography{neurips_2026}

\begin{thebibliography}{10}

\bibitem{besbes2014stochastic}
O.~Besbes, Y.~Gur, and A.~Zeevi.
\newblock Stochastic multi-armed-bandit problem with non-stationary rewards.
\newblock {\em Advances in neural information processing systems}, 27, 2014.

\bibitem{bone2009gis}
C.~Bone and S.~Dragi{\'c}evi{\'c}.
\newblock Gis and intelligent agents for multiobjective natural resource allocation: A reinforcement learning approach.
\newblock {\em Transactions in GIS}, 13(3):253--272, 2009.

\bibitem{busa2017multi}
R.~Busa-Fekete, B.~Sz{\"o}r{\'e}nyi, P.~Weng, and S.~Mannor.
\newblock Multi-objective bandits: Optimizing the generalized gini index.
\newblock In {\em International Conference on Machine Learning}, pages 625--634. PMLR, 2017.

\bibitem{delle2011bounded}
F.~M. Delle~Fave, R.~Stranders, A.~Rogers, and N.~Jennings.
\newblock Bounded decentralised coordination over multiple objectives.
\newblock 2011.

\bibitem{drugan2013designing}
M.~M. Drugan and A.~Nowe.
\newblock Designing multi-objective multi-armed bandits algorithms: A study.
\newblock In {\em The 2013 international joint conference on neural networks (IJCNN)}, pages 1--8. IEEE, 2013.

\bibitem{drugan2014pareto}
M.~M. Drugan, A.~Now{\'e}, and B.~Manderick.
\newblock Pareto upper confidence bounds algorithms: an empirical study.
\newblock In {\em 2014 IEEE Symposium on Adaptive Dynamic Programming and Reinforcement Learning (ADPRL)}, pages 1--8. IEEE, 2014.

\bibitem{garivier2011upper}
A.~Garivier and E.~Moulines.
\newblock On upper-confidence bound policies for switching bandit problems.
\newblock In {\em International conference on algorithmic learning theory}, pages 174--188. Springer, 2011.

\bibitem{grandoni2014utilitarian}
F.~Grandoni, P.~Krysta, S.~Leonardi, and C.~Ventre.
\newblock Utilitarian mechanism design for multiobjective optimization.
\newblock {\em SIAM Journal on Computing}, 43(4):1263--1290, 2014.

\bibitem{hong2021evolutionary}
W.-J. Hong, P.~Yang, and K.~Tang.
\newblock Evolutionary computation for large-scale multi-objective optimization: A decade of progresses.
\newblock {\em International Journal of Automation and Computing}, 18(2):155--169, 2021.

\bibitem{hossain2021fair}
S.~Hossain, E.~Micha, and N.~Shah.
\newblock Fair algorithms for multi-agent multi-armed bandits.
\newblock {\em Advances in Neural Information Processing Systems}, 34:24005--24017, 2021.

\bibitem{huyuk2021multi}
A.~H{\"u}y{\"u}k and C.~Tekin.
\newblock Multi-objective multi-armed bandit with lexicographically ordered and satisficing objectives.
\newblock {\em Machine Learning}, 110(6):1233--1266, 2021.

\bibitem{inja2014queued}
M.~Inja, C.~Kooijman, M.~de~Waard, D.~M. Roijers, and S.~Whiteson.
\newblock Queued pareto local search for multi-objective optimization.
\newblock In {\em International conference on parallel problem solving from nature}, pages 589--599. Springer, 2014.

\bibitem{jones2023efficient}
M.~Jones, H.~Nguyen, and T.~Nguyen.
\newblock An efficient algorithm for fair multi-agent multi-armed bandit with low regret.
\newblock In {\em Proceedings of the AAAI Conference on Artificial Intelligence}, volume~37, pages 8159--8167, 2023.

\bibitem{landgren2016distributed}
P.~Landgren, V.~Srivastava, and N.~E. Leonard.
\newblock On distributed cooperative decision-making in multiarmed bandits.
\newblock In {\em 2016 European Control Conference (ECC)}, pages 243--248. IEEE, 2016.

\bibitem{lee2012multi}
C.-S. Lee.
\newblock Multi-objective game-theory models for conflict analysis in reservoir watershed management.
\newblock {\em Chemosphere}, 87(6):608--613, 2012.

\bibitem{liu2025distributed}
J.~Liu, H.~Qiu, L.~Yang, and M.~Xu.
\newblock Distributed multi-agent bandits over erd$\backslash$h $\{$o$\}$ sr$\backslash$'enyi random networks.
\newblock {\em arXiv preprint arXiv:2510.22811}, 2025.

\bibitem{liu2019emergent}
S.~Liu, G.~Lever, J.~Merel, S.~Tunyasuvunakool, N.~Heess, and T.~Graepel.
\newblock Emergent coordination through competition.
\newblock {\em arXiv preprint arXiv:1902.07151}, 2019.

\bibitem{madani2011monte}
K.~Madani and J.~R. Lund.
\newblock A monte-carlo game theoretic approach for multi-criteria decision making under uncertainty.
\newblock {\em Advances in water resources}, 34(5):607--616, 2011.

\bibitem{mannion2017theoretical}
P.~Mannion, J.~Duggan, and E.~Howley.
\newblock A theoretical and empirical analysis of reward transformations in multi-objective stochastic games.
\newblock 2017.

\bibitem{mouaddib2007towards}
A.-I. Mouaddib, M.~Boussard, and M.~Bouzid.
\newblock Towards a formal framework for multi-objective multiagent planning.
\newblock In {\em Proceedings of the 6th international joint conference on Autonomous agents and multiagent systems}, pages 1--8, 2007.

\bibitem{ruadulescu2020multi}
R.~R{\u{a}}dulescu, P.~Mannion, D.~M. Roijers, and A.~Now{\'e}.
\newblock Multi-objective multi-agent decision making: a utility-based analysis and survey.
\newblock {\em Autonomous Agents and Multi-Agent Systems}, 34(1):10, 2020.

\bibitem{ramos2019budged}
G.~D.~O. Ramos, R.~Radulescu, and A.~Now{\'e}.
\newblock A budged-balanced tolling scheme for efficient equilibria under heterogeneous preferences.
\newblock In {\em Proceedings of the Adaptive and Learning Agents Workshop 2019 (ALA-19) at AAMAS}, 2019.

\bibitem{sankararaman2019social}
A.~Sankararaman, A.~Ganesh, and S.~Shakkottai.
\newblock Social learning in multi agent multi armed bandits.
\newblock {\em Proceedings of the ACM on Measurement and Analysis of Computing Systems}, 3(3):1--35, 2019.

\bibitem{shi2026communication}
M.~Shi.
\newblock Communication-corruption coupling and verification in cooperative multi-objective bandits.
\newblock {\em arXiv preprint arXiv:2601.11924}, 2026.

\bibitem{taylor2014accelerating}
A.~Taylor, I.~Dusparic, E.~Galvan-Lopez, S.~Clarke, and V.~Cahill.
\newblock Accelerating learning in multi-objective systems through transfer learning.
\newblock In {\em 2014 international joint conference on neural networks (IJCNN)}, pages 2298--2305. IEEE, 2014.

\bibitem{wang2022achieving}
X.~Wang, L.~Yang, Y.-Z.~J. Chen, X.~Liu, M.~Hajiesmaili, D.~Towsley, and J.~C. Lui.
\newblock Achieving near-optimal individual regret \& low communications in multi-agent bandits.
\newblock In {\em The Eleventh International Conference on Learning Representations}, 2022.

\bibitem{xu2023decentralized}
M.~Xu and D.~Klabjan.
\newblock Decentralized randomly distributed multi-agent multi-armed bandit with heterogeneous rewards.
\newblock {\em Advances in Neural Information Processing Systems}, 36:74799--74855, 2023.

\bibitem{xu2025heterogeneous}
M.~Xu, L.~Shan, F.~Ghaffari, X.~Wang, X.~Liu, and M.~Hajiesmaili.
\newblock Heterogeneous multi-agent multi-armed bandit on stochastic block models.
\newblock {\em Proceedings of the ACM on Measurement and Analysis of Computing Systems}, 9(3):1--59, 2025.

\bibitem{yahyaa2014annealing}
S.~Q. Yahyaa, M.~M. Drugan, and B.~Manderick.
\newblock Annealing-pareto multi-objective multi-armed bandit algorithm.
\newblock In {\em 2014 IEEE Symposium on Adaptive Dynamic Programming and Reinforcement Learning (ADPRL)}, pages 1--8. IEEE, 2014.

\bibitem{zhang2024no}
M.~Zhang, R.~Deo-Campo~Vuong, and H.~Luo.
\newblock No-regret learning for fair multi-agent social welfare optimization.
\newblock {\em Advances in Neural Information Processing Systems}, 37:57671--57700, 2024.

\bibitem{zhu2020distributed}
J.~Zhu, R.~Sandhu, and J.~Liu.
\newblock A distributed algorithm for sequential decision making in multi-armed bandit with homogeneous rewards.
\newblock In {\em 2020 59th IEEE Conference on Decision and Control (CDC)}, pages 3078--3083. IEEE, 2020.

\end{thebibliography}


\newpage

\appendix

\section{Technical appendices and supplementary material}

\subsection{Related Work}\label{app:related_work}

\textbf{MA-MAB} The study of multi-agent multi-armed bandits (MA-MAB) with a fixed set of agents has gained significant attention in distributed learning systems \citep{landgren2016distributed, zhu2020distributed, xu2023decentralized, wang2022achieving, sankararaman2019social, xu2025heterogeneous}. A central distinction in this literature is between homogeneous settings, where all agents share identical reward distributions for each arm \citep{landgren2016distributed, zhu2020distributed} and heterogeneous settings, where reward distributions vary across agents \citep{xu2023decentralized}. While homogeneous models are more tractable and better understood, heterogeneous settings introduce additional challenges due to conflicting observations and agent-specific variability.

Another key dimension is the communication structure among agents. In centralized approaches, a global coordinator aggregates information from all agents, whereas in decentralized settings, agents communicate locally over a network. Within decentralized MA-MAB, different graph models have been studied, including fully connected networks \citep{landgren2016distributed}, connected deterministic graphs \citep{wang2022achieving, sankararaman2019social}, random graphs \citep{xu2023decentralized}, and structured models such as stochastic block graphs \cite{xu2025heterogeneous}. These choices significantly impact learning performance, as they govern how efficiently information propagates across agents.

Despite this progress, existing work typically studies these aspects in isolation and does not fully address the combined challenges of heterogeneous rewards and time-varying decentralized communication, particularly in multi-objective settings. Our work builds on this literature by integrating these components within a unified framework.

A closely related line of work studies multi-agent bandits with fairness objectives, particularly through Nash Social Welfare (NSW) regret. Prior work \citep{hossain2021fair, jones2023efficient} considers a setting with a utility matrix specifying rewards for each agent under homogeneous rewards and centralized communication, while subsequent extensions address non-stationary utilities \citep{zhang2024no}. These approaches operate on scalar utilities, where each entry of the utility matrix directly represents an agent’s reward for an arm. In contrast, our formulation incorporates multi-dimensional rewards, where utilities are derived as linear combinations of reward components weighted by agent-specific preference vectors, building on \citep{hossain2021fair}.

\textbf{MO-MAB} Multi-objective multi-armed bandits (MO-MAB) extend the classical bandit framework by replacing scalar rewards with vector-valued rewards, which introduces significant challenges in both learning and evaluation. A natural approach is to extend confidence-bound methods to this setting. In particular, \textsc{Pareto UCB1} \citep{drugan2013designing} maintains coordinate-wise confidence bounds and identifies an estimated Pareto front rather than a single optimal arm, from which actions are selected. This approach achieves logarithmic Pareto regret using multi-dimensional concentration inequalities, with subsequent refinements improving performance \citep{drugan2014pareto}. However, such methods rely on maintaining and exploring the Pareto front, whose size and complexity can grow with the reward dimension, making efficient learning increasingly difficult in high-dimensional settings. Other variants, such as PF-LEX \citep{huyuk2021multi} incorporate alternative preference structures like lexicographic ordering, though they similarly operate within a Pareto-based framework. 

Despite these advances, existing approaches remain largely confined to single-agent settings and focus on Pareto optimality as the primary notion of performance. A complementary line of work from multi-objective optimization (MOO) considers algorithms such as evolutionary methods \citep{hong2021evolutionary} and Bayesian optimization techniques (e.g., annealing Pareto knowledge gradient \citep{yahyaa2014annealing}), which have demonstrated strong empirical performance in navigating trade-offs across objectives. However, these methods typically lack regret guarantees, limiting their applicability in online learning contexts. Another common strategy is to apply scalarization, converting the vector-valued problem into a scalar bandit problem via linear or Chebyshev functions. While computationally appealing, such approaches obscure the underlying multi-objective structure and depend heavily on the choice of preference weights. Related work optimizing alternative welfare criteria, such as the General Gini Index (GGI) \citep{busa2017multi} similarly evaluates performance relative to a fixed scalarization rather than directly addressing multi-objective trade-offs. 

In contrast to existing MO-MAB literature, which primarily focuses on Pareto-based optimality, we adopt a utility-based perspective for evaluating vector-valued rewards. By modeling agent-specific preferences through preference vectors, we map multi-dimensional rewards to utilities and move beyond Pareto front identification toward optimizing aggregated objectives. This formulation naturally enables the incorporation of social welfare criteria, such as Nash Social Welfare, linking multi-objective bandits with fairness-aware optimization.

\textbf{MO-MA-MAB} Multi-objective multi-agent decision making has been studied in the context of multi-objective partially observable stochastic games (MOPOSGs) \citep{ruadulescu2020multi}. These provide a general framework that can be specialized along key dimensions—observability, the form of the reward function, and the temporal structure of interactions—allowing a wide range of problem settings to be identified. The presence of multiple objectives complicates the notion of optimality, as agents may evaluate outcomes according to different utility functions. A useful way to structure these problems is through the distinction between rewards and utilities. Rewards may be team-based \citep{bone2009gis, delle2011bounded, inja2014queued, mannion2017theoretical, lee2012multi}, where all agents observe a common vector-valued reward, or individual \citep{madani2011monte, taylor2014accelerating}, where rewards differ across agents. Utilities can be shared (team utilities) \citep{bone2009gis, delle2011bounded, inja2014queued, mannion2017theoretical}, agent-specific (individual utilities) \citep{lee2012multi, madani2011monte, taylor2014accelerating}, or aggregated to capture fairness considerations (social choice) \citep{grandoni2014utilitarian, liu2019emergent, mouaddib2007towards, ramos2019budged}. Different solution concepts arise depending on this formulation: coverage sets for team utilities, equilibrium and stability notions for individual utilities, and mechanism design approaches for social choice settings.

A recent line of work considers cooperative multi-objective bandits under adversarial corruption \citep{shi2026communication}. In this setting, agents operate under homogeneous rewards with centralized communication, and performance is measured via regret defined through an $L$-Lipschitz scalarization of vector-valued rewards. The key challenge is to ensure reliable learning when observations may be corrupted by an adversary. Specifically, a bounded number of observations can be verified as uncorrupted, controlled by a budget $v$, while the remaining observations may be perturbed by a corruption vector with total magnitude constrained by $\tau$. The authors study three communication strategies—raw sample sharing, sufficient statistic sharing, and recommendation-only sharing—and characterize how corruption impacts regret in each case. In particular, sharing raw samples propagates corrupted information and leads to regret scaling with both the corruption budget and the number of agents, whereas the latter two strategies mitigate this effect and depend only on the corruption level.

\subsection{Regret Bound for Pareto UCB1 Gossip}\label{proof:mo}

We use the structure of the proof of \textsc{Pareto UCB1}\citep{drugan2013designing} using ideas from the proof of \textsc{Gossip Successive Elimination}\citep{liu2025distributed} to bound the deviation between the global average and the centered mean vector. We first bound
$$\Pr [z_{i,k}(t+1)\nprec\mu_k +\epsilon\mathbf{1} ]\leq\Pr\left[ z_{i,k}(t+1)\nprec\tilde{\mu}_k (t) +\frac{\epsilon}{2}\mathbf{1}\right]+\Pr\left[\tilde{\mu}_k (t)\nprec\mu_k +\frac{\epsilon}{2}\mathbf{1}\right]$$
Note that we use the decomposition into consensus and estimation error like in \citep{liu2025distributed}. The intermediate $\tilde{\mu}_k (t)=\frac{1}{n}\sum_{i=1}^n\hat{\mu}_{i,k} (t)$ is the average of the local averages. Then $\tilde{\mu}_k (t)$ is a weighted average of independent random variables. We can bound the estimation error using Chernoff and union bound. By Chernoff's bound, for a weighted sum of random variables
$$\Pr [S>\mathbb{E} [S]+\epsilon ]\leq\exp\left(\frac{-2\epsilon^2}{\sum_j a_j^2}\right)$$
Then we can represent $\tilde{\mu}_k^d (t)$ as
$$\tilde{\mu}_k^d (t)=\frac{1}{n}\sum_{i=1}^n\hat{X}_{i,k}^d (t)=\frac{1}{n}\sum_{i=1}^n\frac{1}{T_{i,k}(t)}\sum_{\tau =1}^t\ind_{\{ A_i(\tau )=k\}} X_{i,k}^d(\tau )=\sum_{i=1}^n\sum_{\tau =1}^t\frac{1}{nT_{i,k}(t)}\ind_{\{ A_i(\tau )=k\}} X_{i,k}(\tau )$$
In this case
$$\sum_j a_j^2=\sum_{i=1}^n\sum_{\tau =1}^{T_{i,k}(t)}\frac{1}{(nT_{i,k}(t))^2} =\sum_{i=1}^n\frac{1}{n^2T_{i,k}(t)}\leq\frac{1}{n}\sum_{i=1}^n\frac{1}{T_{i,k}(t)}$$
For $d\in [D]$
$$\Pr\left[\tilde{\mu}_k^d (t)>\mu_k^d +\frac{\epsilon}{2}\right]\leq\exp\left(\frac{-\frac{1}{2} n\epsilon^2}{\sum_{i=1}^n\frac{1}{T_{i,k}(t)}}\right)\leq\exp\left(\frac{-\frac{1}{2} n\epsilon^2}{\frac{n}{s_k(t)}}\right) =\exp\left( -\frac{1}{2} s_k(t)\epsilon^2\right)$$
where $s_k=\min_{i\in [n]} T_{i,k}(t)$. Therefore, by union bound the estimation error
$$\Pr\left[\tilde{\mu}_k (t)\nprec\mu_k +\frac{\epsilon}{2}\mathbf{1}\right] =\Pr \left[\cup_{d=1}^D\tilde{\mu}_k^d (t)>\mu_k^d +\frac{\epsilon}{2}\right]\leq D\exp\left( -\frac{1}{2} s_k(t)\epsilon^2\right)$$
To bound consensus error like in \citep{liu2025distributed} we find a recurrence, viewing it as a orthogonal projection from the average of the local averages. To achieve this, we unroll and using the row stochasticity property of $W_t$. Departing from \citep{liu2025distributed}, we vectorize across agents $i$ and dimensions $d$. $z_k(t+1)\in\mathbb{R}^{n\times D}$ is the global average at time $t+1$ and $\tilde{\mu}_k (t)\in\mathbb{R}^{n\times D}$ is the average of the local averages at time $t$. We let $e_k(t+1)=z_k(t+1)-\tilde{\mu}_k (t)$ be the consensus error at time $t+1$. Lastly suppose $u_k(t)=\hat{\mu}_k (t)-\hat{\mu}_k (t)$ is the local update at $t$.
\begin{align*}
    e_k(t+1) & =z_k(t+1)-\tilde{\mu}_k (t) \\
    & =z_k(t+1)-\frac{1}{n}\mathbf{1}\mathbf{1}^\top\hat{\mu}_k (t) \\
    & =W_tz_k(t)+u_k(t)-\frac{1}{n}\mathbf{1}\mathbf{1}^\top\hat{\mu}_k (t) \\
    & =W_tz_k(t)+u_k(t)-\frac{1}{n}\mathbf{1}\mathbf{1}^\top\hat{\mu}_k (t-1)-\frac{1}{n}\mathbf{1}\mathbf{1}^\top u_k(t) \\
    & =\left( W_tz_k(t)-\frac{1}{n}\mathbf{1}\mathbf{1}^\top\hat{\mu}_k (t-1)\right) +\left( I-\frac{1}{n}\mathbf{1}\mathbf{1}^\top\right) u_k(t) \\
    & =W_te_k(t)+\left( I-\frac{1}{n}\mathbf{1}\mathbf{1}^\top \right) u_k(t)
\end{align*}
Unrolling we have
$$e_k(t)=\sum_{\tau =1}^{t-1}\left(\prod_{s=\tau +1}^{t-1} W_s\right)\left( I-\frac{1}{n}\mathbf{1}\mathbf{1}^\top\right) u_k(\tau )$$
Next, we bound the Frobenius norm of $e_k(t)$.
\begin{align*}
    \norm{e_k(t)}_F^2 & =\sum_{d=1}^D\norm{\sum_{\tau =1}^{t-1}\left(\prod_{s=\tau +1}^{t-1} W_s\right)\left( I-\frac{1}{n}\mathbf{1}\mathbf{1}^\top\right) u_k^d(\tau )}_2^2 \\
    &\leq\sum_{d=1}^D\sum_{\tau =1}^{t-1}\norm{\left(\prod_{s=\tau +1}^{t-1} W_s\right)\left( I-\frac{1}{n}\mathbf{1}\mathbf{1}^\top\right) u_k^d(\tau )}_2^2 \\
    &\leq\sum_{d=1}^D\sum_{\tau =1}^{t-1}\norm{\left(\prod_{s=\tau +1}^{t-1} W_s\right)\left( I-\frac{1}{n}\mathbf{1}\mathbf{1}^\top\right)}_{op}^2\norm{u_k^d(\tau )}_2^2 \\
    & =\sum_{\tau =1}^{t-1}\norm{\left(\prod_{s=\tau +1}^{t-1} W_s\right)\left( I-\frac{1}{n}\mathbf{1}\mathbf{1}^\top\right)}_{op}^2\sum_{d=1}^D\norm{u_k^d(\tau )}_2^2 \\
    & =\sum_{\tau =1}^{t-1}\norm{\left(\prod_{s=\tau +1}^{t-1} W_s\right)\left( I-\frac{1}{n}\mathbf{1}\mathbf{1}^\top\right)}_{op}^2\norm{u_k(\tau )}_F^2
\end{align*}
Referencing preliminary fact B3 in \citep{liu2025distributed}, if $W_t$ is double stochastic, symmetric, and iid, for all $v\in V$ and $s,t\in\N$ with $s<t$
$$\Pr\left[\norm{W_t\dots W_{s+1}\frac{1}{n}\mathbf{1}}\geq\delta\right]\leq\frac{\lambda_2 (\E [W^2])^{t-s}}{\delta^2}$$
Let $A_{t,\tau}=W_{t-1}\dots W_{\tau +1}(I-\frac{1}{n}\mathbf{1}\mathbf{1}^\top )$. Due to the row stochasticity of $W_t$
$$A_{t,\tau} e_i=W_{t-1}\dots W_{\tau +1}\left( e_i-\frac{1}{n}\mathbf{1}\mathbf{1}^\top e_i\right) =W_{t-1}\dots W_{\tau +1}\left( e_i-\frac{1}{n}\mathbf{1}\right) =W_{t-1}\dots W_{\tau +1}e_i-\frac{1}{n}\mathbf{1}$$
Since $\norm{A_{t,\tau}}_{op}\leq\sqrt{\sum_{i=1}^n\norm{Ae_i}_2^2}$,
\begin{align*}
    \Pr [\norm{A_{t,\tau}}_{op}\geq\delta ] & \leq \Pr\left[\sum_{i=1}^n\norm{A_{t,\tau} e_i}_2^2\geq\delta^2\right] \\
    & \leq\sum_{i=1}^n\Pr\left[\norm{A_{t,\tau} e_i}_2^2\geq\frac{\delta}{n}\right] =\sum_{i=1}^n\Pr\left[\norm{A_{t,\tau} e_i}_2\geq\frac{\delta}{\sqrt{n}}\right] \\
    & \leq\sum_{i=1}^n\frac{n\lambda_2 (\E [W^2])^{t-\tau -1}}{\delta^2} =\frac{n^2\lambda_2 (\E [W^2])^{t-\tau -1}}{\delta^2}
\end{align*}
Let $\rho =\lambda_2 (\E [W^2])$ where $\rho <1$. Setting $\delta =\rho^{(t-s-1)/2}t$, the bound holds with probability $\frac{n^2}{t^2}$. We use the union bound to show 
$$\Pr [\exists\tau <t:\norm{A_{t,\tau}}_{op}\geq\rho^{(t-s-1)/2} t]\leq\sum_{t=1}^{t-1}\Pr [\norm{A_{t,\tau}}_{op}\geq\rho^{(t-s-1)/2} t]=(t-1)\frac{n^2}{t^2}\leq\frac{n^2}{t}$$
As a result,
$$\Pr [\forall\tau <t:\norm{A_{t,\tau}}_{op}\leq\rho^{(t-s-1)/2} t]\geq 1-\frac{n^2}{t}$$
Now we bound the second term in the product $\norm{u_k(\tau )}_F^2$. In the worse case scenario, we have that arm $k$ is pulled at $\tau$. Recall that
$$\hat{\mu}_{i,k} (\tau )=\frac{T_{i,k}(\tau -1)\hat{\mu}_{i,k} (\tau -1)+X_{i,k}(\tau )}{T_{i,k}(\tau -1)+1}$$
Then
\begin{align*}
    \hat{\mu}_{i,k} (\tau )-\hat{\mu}_{i,k} (\tau -1) & =\frac{T_{i,k}(\tau -1)\hat{\mu}_{i,k} (\tau -1)+X_{i,k}(\tau )-T_{i,k}(\tau )\hat{\mu}_{i,k} (\tau -1)}{T_{i,k} (\tau )} \\
    & =\frac{X_{i,k}(\tau )-\hat{\mu}_{i,k} (\tau -1)}{T_{i,k} (\tau )}\leq\frac{1}{T_{i,k} (\tau )}
\end{align*}
as rewards are in the range $[0,1]$. Summing over agents $i\in [n]$
$$\norm{u_k(\tau )}_F^2=\sum_{i=1}^n (\hat{\mu}_{i,k} (\tau )-\hat{\mu}_{i,k} (\tau -1))^2\leq\sum_{i=1}^n\frac{1}{T_{i,k}(\tau )^2}\leq\frac{n}{\tau^2}$$
Overall,
\begin{align*}
    \norm{e_k(t)}_F & \leq t\sqrt{n}\sum_{\tau =1}^{t-1}\frac{\rho^{(t-\tau -1)/2}}{\tau} \\
    & =t\sqrt{n}\sum_{j=0}^{t-2}\frac{\rho^{j/2}}{t-j-1} \\
    & \leq (t\sqrt{n} )\frac{1}{t}\sum_{j=0}^{t-2}\rho^{j/2}\leq\sqrt{n}\sum_{j=0}^\infty\rho^{j/2} =\frac{\sqrt{n}}{1-\sqrt{\rho}}
\end{align*}
with probability at least $1-\frac{n^2}{t}$. If the Frobenius norm of the consensus error matrix is bounded, each of the entries in the matrix is bounded.
$$\Pr\left[\cap_{i=1}^n\cap_{d=1}^D\abs{z_{i,k}^d(t)-\tilde{\mu}_k^d}\leq\frac{\sqrt{n}}{1-\sqrt{\rho}}\right]\geq 1-\frac{n^2}{t}$$
Then for all agents $i\in [n]$, setting $\epsilon\geq\frac{2\sqrt{n}}{1-\sqrt{p}}$
$$\Pr\left[ z_{i,k}(t+1)\nprec\tilde{\mu}_k (t) +\frac{\epsilon}{2}\mathbf{1}\right]\leq\Pr\left[\cup_{d=1}^D z_{i,k}^d(t)>\tilde{\mu}_k^d (t)+\frac{\sqrt{n}}{1-\sqrt{\rho}}\right]\leq\frac{n^2}{t}$$
Combining estimation and consensus bounds
$$\Pr [z_{i,k}(t+1)\nprec\mu_k +\epsilon\mathbf{1} ]\leq D\exp\left( -\frac{1}{2} s_k(t)\epsilon^2\right) +\frac{n^2}{t}$$
Using the same argument with $\epsilon\geq\frac{2\sqrt{n}}{1-\sqrt{p}}$
$$\Pr [z_{i,k}(t+1)\nsucc\mu_k -\epsilon\mathbf{1} ]\leq D\exp\left( -\frac{1}{2} s_k(t)\epsilon^2\right) +\frac{n^2}{t}$$
Recall that $s_k=\min_{i\in [n]} T_{i,k}(t)$. The Pareto UCB1 Gossip exploration term is given by $c_{i,k}(t)=\sqrt{\frac{8\log (t\sqrt[4]{D\abs{\mathcal{A}^*}} )}{T_{i,k}(t)}} +\frac{2\sqrt{n}}{1-\sqrt{p}}$. Define $d_k(t)=\sqrt{\frac{8\log (t\sqrt[4]{D\abs{\mathcal{A}^*}} )}{s_k(t)}} +\frac{2\sqrt{n}}{1-\sqrt{p}}$ so that $d_k(t)\geq c_{i,k}(t)$. Substituting $\epsilon =d_k(t)$ into the bounds
$$\Pr [z_{i,k}(t+1)\nprec\mu_k +d_k(t)\mathbf{1} ]\leq D\exp\left( -\frac{1}{2} s_k(t)\left(\frac{8\log (t\sqrt[4]{D\abs{\mathcal{A}^*}} )}{s_k(t)}\right)\right) +\frac{n^2}{t} =D(t\sqrt[4]{D\abs{\mathcal{A}^*}} )^{-4}+\frac{n^2}{t}=\frac{1}{\abs{\mathcal{A}^*}t^4} +\frac{n^2}{t}$$
Equivalently, we have
$$\Pr [z_{i,k}(t+1)\nsucc\mu_k -d_k(t)\mathbf{1} ]\leq\frac{1}{\abs{\mathcal{A}^*}t^4} +\frac{n^2}{t}$$
The goal is to bound the counter of a suboptimal arm $k$ for a particular agent $i$. Let $l>0$ be the number of times arm $k$ is pulled during a ``warm-up period".
\begin{align*}
    T_{i,k}(T) & =1+\sum_{t=K+1}^T\ind_{\{ A_i(t)=k\}} \\
    & \leq l+\sum_{t=K+1}^T\ind_{\{ A_i(t)=k,T_{i,k}(t-1)\geq l\}} \\
    & \leq l+\sum_{t=K+1}^T\sum_{h=1}^{\abs{\mathcal{A}^*}}\ind_{\{ z_{i,k}(t+1)+c_{i,k}(t)\nprec z_{i,h}(t+1)+c_{i,h}(t),T_{i,k}(t-1)\geq l\}} \\
    & \leq l+\sum_{t=K+1}^T\sum_{s_k=l}^T\sum_{s_h=1}^T\sum_{h=1}^{\abs{\mathcal{A}^*}}\ind_{\{z_{i,k}(t+1)+c_{i,k}(s_k)\nprec z_{i,h}(t+1)+c_{i,h}(s_h),T_{i,k}(t-1)=s_k,T_{i,h}(t-1)=s_h\}}
\end{align*}
where $c_{i,k}(s_k)$ is the explore term for arm $k$ after it is pulled $s_k$ times and $c_{i,h}(s_h)$ is the explore term for arm $h$ after it is pulled $s_h$ times. Note that the global averages cannot be indexed by counters but the explore terms can. For the last inequality fixing a Pareto arm $h$, at a time $t$ the indicator is upper bounded by summing over the possible counter values. The counter $s_k$ for arm $k$ can take values from $l$ to $T$. The counter $s_h$ for arm $h$ can take values from $1$ to $T$. If UCB of arm $k$ with count $s_k$ is not dominated by UCB of arm $h$ with count $s_k$, at least one of the following conditions hold.
$$z_{i,h}(t+1)\nsucc\mu_h -c_{i,h}(s_h)\mathbf{1} ;z_{i,k}(t+1)\nprec\mu_k +c_{i,k}(s_k)\mathbf{1} ;\mu_h\nsucc\mu_k +2c_{i,k}(s_k)\mathbf{1}$$
After an interval of time, it happens that statement 3 does not hold. This happens when $2c_{i,k}(s_k)\leq\Delta_k$. This implies that
$$2\left(\sqrt{\frac{8\log (t\sqrt[4]{D\abs{\mathcal{A}^*}} )}{s_k}} +\frac{2\sqrt{n}}{1-\sqrt{p}}\right)\leq\Delta_k$$
Using the stronger inequality $2\sqrt{\frac{8\log (t\sqrt[4]{D\abs{\mathcal{A}^*}} )}{s_k}}\leq\Delta_k$, when $s_k\geq\frac{32\log (t\sqrt[4]{D\abs{\mathcal{A}^*}})}{\Delta_k^2}$ statement 3 does not hold. Set the warm-up period $l=\ceil{\frac{32\log (t\sqrt[4]{D\abs{\mathcal{A}^*}})}{\Delta_k^2}}$. After this warm-up period if arm $k$ is not dominated, statement 1 or 2 must hold. Taking expectations on both sides
\begin{align*}
    \E [T_{i,k}(t)] & \leq l+\sum_{t=K+1}^T\sum_{s_k=l}^T\sum_{s_h=1}^T\sum_{h=1}^{\abs{\mathcal{A}^*}}\Pr [z_{i,k}(t+1)+c_{i,k}(s_k)\nprec z_{i,h}(t+1)+c_{i,h}(s_h),T_{i,k}(t-1)=s_k,T_{i,h}(t-1)=s_h] \\
    \begin{split}
        & \leq l+\sum_{t=K+1}^T\sum_{s_k=l}^T\sum_{s_h=1}^T\sum_{h=1}^{\abs{\mathcal{A}^*}}\Pr [z_{i,h}(t+1)\nsucc\mu_h -c_{i,h}(s_h)\mathbf{1} ,T_{i,k}(t-1)=s_k,T_{i,h}(t-1)=s_h] \\
        &\qquad +\sum_{t=K+1}^T\sum_{s_k=l}^T\sum_{s_h=1}^T\sum_{h=1}^{\abs{\mathcal{A}^*}}\Pr [z_{i,k}(t+1)\nprec\mu_k +c_{i,k}(s_k)\mathbf{1} ,T_{i,k}(t-1)=s_k,T_{i,h}(t-1)=s_h]
    \end{split} \\
    \begin{split}
        & =l+\sum_{h=1}^{\abs{\mathcal{A}^*}}\sum_{t=K+1}^T\sum_{s_h=1}^T\Pr [z_{i,h}(t+1)\nsucc\mu_h -c_{i,h}(s_h)\mathbf{1} ,T_{i,h}(t-1)=s_h] \\
        &\qquad +\sum_{h=1}^{\abs{\mathcal{A}^*}}\sum_{t=K+1}^T\sum_{s_k=l}^T\Pr [z_{i,k}(t+1)\nprec\mu_k +c_{i,k}(s_k)\mathbf{1} ,T_{i,k}(t-1)=s_k]
    \end{split} \\
    \begin{split}
        & =l+\sum_{h=1}^{\abs{\mathcal{A}^*}}\sum_{t=K+1}^T\sum_{s_h=1}^T\Pr [z_{i,h}(t+1)\nsucc\mu_h -c_{i,h}(s_h)\mathbf{1} |T_{i,h}(t-1)=s_h]\Pr [T_{i,h}(t-1)=s_h] \\
        &\qquad +\sum_{h=1}^{\abs{\mathcal{A}^*}}\sum_{t=K+1}^T\sum_{s_k=l}^T\Pr [z_{i,k}(t+1)\nprec\mu_k +c_{i,k}(s_k)\mathbf{1} |T_{i,k}(t-1)=s_k]\Pr [T_{i,k}(t-1)=s_k]
    \end{split} \\
    \begin{split}
        & \leq l+\sum_{h=1}^{\abs{\mathcal{A}^*}}\sum_{t=K+1}^T\sum_{s_h=1}^T\left(\frac{1}{\abs{\mathcal{A}^*}t^4} +\frac{n^2}{t}\right)\Pr [T_{i,h}(t-1)=s_h] \\
        &\qquad +\sum_{h=1}^{\abs{\mathcal{A}^*}}\sum_{t=K+1}^T\sum_{s_k=l}^T\left(\frac{1}{\abs{\mathcal{A}^*}t^4} +\frac{n^2}{t}\right)\Pr [T_{i,k}(t-1)=s_k]
    \end{split} \\
    & =l+\sum_{h=1}^{\abs{\mathcal{A}^*}}\sum_{t=K+1}^T 2\left(\frac{1}{\abs{\mathcal{A}^*}t^4} +\frac{n^2}{t}\right) =l+2\sum_{t=K+1}^T\frac{1}{t^4} +\frac{\abs{\mathcal{A}^*}n^2}{t}\leq l+2\sum_{t=K+1}^\infty\frac{1}{t^4} +2\abs{\mathcal{A}^*} n^2\sum_{t=K+1}^T\frac{1}{t} \\
    & \leq l+\frac{\pi^4}{45} +2\abs{\mathcal{A}^*} n^2(\log T+1)
\end{align*}
The Pareto regret of a policy $\pi$ at time $t$ for a single agent $i$ is bounded by
$$\sum_{k\notin\mathcal{A}^*}\frac{32\log (T\sqrt[4]{D\abs{\mathcal{A}^*}})}{\Delta_k} +\left(1+\frac{\pi^4}{45} +2\abs{\mathcal{A}^*} n^2(\log T+1)\right)\sum_{k\in\mathcal{A}^*}\Delta_k$$
As a result, the global Pareto regret of a policy $\pi$ at time $t$
$$\text{Reg}_T(\pi )\leq n\sum_{k\notin\mathcal{A}^*}\frac{32\log (T\sqrt[4]{D\abs{\mathcal{A}^*}})}{\Delta_k} +n\left(1+\frac{\pi^4}{45} +2\abs{\mathcal{A}^*} n^2(\log T+1)\right)\sum_{k\in\mathcal{A}^*}\Delta_k$$

\subsection{Regret Bound for Simulated NSW UCB Gossip}\label{app:proof_NSW}

NSW regret over agents $i\in [N]$ over time steps $t\in [T]$ is given by 
$$R_T=\sum_{i=1}^N\sum_{t=1}^T\text{NSW} (p^*,\bar{\mu} )-\text{NSW} (p_i^t,\bar{\mu} )$$
where $\bar{\mu}_{j,k} =\frac{1}{N}\sum_{i=1}^N \mu_{i,j,k}^*$. First, we will bound the regret incurred during the first $K$ time steps. 
$$\sum_{i=1}^N\sum_{t=1}^K\text{NSW} (p^*,\bar{\mu} )-\text{NSW} (p_i^t,\bar{\mu} )\leq NK\text{NSW} (p^*,\bar{\mu} )=NK$$
since rewards are restricted to the interval $[0,1]$. It suffices to consider regret incurred post warm up period. We will bound the deviation between the global average and the centered arm mean using the decomposition
$$z_{i,j,k}(t)-\bar{\mu}_{j,k} =(z_{i,j,k}(t)-\bar{z}_{j,k} (t))+(\bar{z}_{j,k} (t)-\bar{\mu}_{j,k} )$$
Note that $\bar{z}_{j,k} (t)=\frac{1}{N}\sum_{i=1}^N z_{i,j,k}(t)$. The first term in the decomposition represents the consensus error and the second term represents the sampling error. We will bound each term separately with a series of lemmas. In Lemma 1, we will draw a meaningful connection between the global averages and local averages. In particular, we show that for all $j\in [N]$, $k\in [K]$, and $t\in [T]$
$$\bar{z}_{j,k} (t)=\frac{1}{N}\sum_{i=1}^N\hat{\mu}_{i,j,k} (t-1)$$
Consequently, we have that $z_{j,k}(t)$ is an unbiased estimator of $\mu_{j,k}$. That is, $\E [z_{j,k}(t)]=\frac{1}{N}\sum_{i=1}^N\E [\hat{\mu}_{i,j,k} (t-1)]=\bar{\mu}_{j,k}$. Suppose that $z_{j,k}(t)$ and $\Delta\hat{\mu}_{j,k} (t)=\hat{\mu}_{j,k} (t)-\hat{\mu}_{j,k} (t)$ are vectors in $\R^N$ defined as expected. We can rewrite the gossip update as
$$z_{j,k}(t+1)=W_tz_{j,k}(t)+\Delta\hat{\mu}_{j,k} (t)$$
We can vectorize the centered global average as $\bar{z}_{j,k} (t)=\frac{1}{N}\ind^\top z_{j,k}(t)$ making
$$\bar{z}_{j,k} (t+1)=\frac{1}{N}\ind^\top W_tz_{j,k}(t)+\frac{1}{N}\ind^\top\Delta\hat{\mu}_{j,k} (t)$$
Since $W_t$ is doubly stochastic,
$$\bar{z}_{j,k} (t+1)=\frac{1}{N}\ind^\top z_{j,k}(t)+\frac{1}{N}\ind^\top\Delta\hat{\mu}_{j,k} (t)=\bar{z}_{j,k} (t)+\frac{1}{N}\sum_{i=1}^N\Delta\hat{\mu}_{i,j,k} (t)$$
Unrolling the recursion we have
$$\bar{z}_{j,k} (t)=\bar{z}_{j,k} (1)+\frac{1}{N}\sum_{s=1}^{t-1}\sum_{i=1}^N\Delta\hat{\mu}_{i,j,k} (s)$$
By definition we also have that
$$\hat{\mu}_{i,j,k} (t-1)=\hat{\mu}_{i,j,k} (0)+\sum_{s=1}^{t-1}\Delta\hat{\mu}_{i,j,k} (s)$$
making
$$\frac{1}{N}\sum_{i=1}^N\hat{\mu}_{i,j,k} (t-1)=\frac{1}{N}\sum_{i=1}^N\hat{\mu}_{i,j,k} (0)+\frac{1}{N}\sum_{s=1}^{t-1}\sum_{i=1}^N\Delta\hat{\mu}_{i,j,k} (s)$$
Since $z_{i,j,k}(1)=\hat{\mu}_{i,j,k} (0)=0$, we know that $\bar{z}_{j,k} (1)=\frac{1}{N}\sum_{i=1}^N\hat{\mu}_{i,j,k} (0)$. This allows us to conclude $\bar{z}_{j,k} (t)=\frac{1}{N}\sum_{i=1}^N\hat{\mu}_{i,j,k} (t-1)$. In Lemma 2, we bound the consensus error for all $i,j\in [N]$, $k\in [K]$, and $t\in [T]$. Assume that
$$\prod_{\tau =s}^{t-1} W_\tau=\frac{1}{N}\ind\ind^\top +E_{s:t}\text{ where }\norm{E_{s:t}}_2\leq\rho^{t-s}<1$$
We can unroll the gossip update for $z_{j,k}(t+1)$.
\begin{align*}
    z_{j,k}(1) & =0 \\
    z_{j,k}(2) & =W_1z_{j,k}(1)+\Delta\hat{\mu}_{j,k} (1)=\Delta\hat{\mu}_{j,k} (1) \\
    z_{j,k}(3) & =W_2z_{j,k}(2)+\Delta\hat{\mu}_{j,k} (2)=W_2\Delta\hat{\mu}_{j,k} (1)+\Delta\hat{\mu}_{j,k} (2) \\
    z_{j,k}(4) & =W_3z_{j,k}(3)+\Delta\hat{\mu}_{j,k} (3)=W_3W_2\Delta\hat{\mu}_{j,k} (1)+W_3\Delta\hat{\mu}_{j,k} (2)+\Delta\hat{\mu}_{j,k} (3) \\
    z_{j,k}(t) & =\sum_{s=1}^{t-1}\left(\prod_{\tau =s+1}^{t-1} W_t\right)\Delta\hat{\mu}_{j,k} (s)
\end{align*}
We will decompose $\prod_{\tau =s+1}^{t-1} W_\tau=\frac{1}{N}\ind\ind^\top +E_{s+1:t}\text{ where }\norm{W_\tau -\frac{1}{N}\ind\ind^\top}_2\leq\rho$ for every $\tau$. We now claim that 
$$\prod_{\tau =s+1}^{t-1} W_t-\frac{1}{N}\ind\ind^\top =\prod_{\tau =s+1}^{t-1} (W_t-\frac{1}{N}\ind\ind^\top )$$
By induction, for the base case $\tau=s+1$ we have that $W_t-\frac{1}{N}\ind\ind^\top =W_t-\frac{1}{N}\ind\ind^\top$. For the inductive step, we prove that if the claim holds for $\tau =s+m$ for $m\geq 1$, it will hold for $\tau =s+m+1$. Letting $P_m=\prod_{\tau =s+1}^{s+m} W_t$,
$$\prod_{\tau =s+1}^{s+m+1} W_t-\frac{1}{N}\ind\ind^\top =P_mW_{s+m+1}-P_m\frac{1}{N}\ind\ind^\top +P_m\frac{1}{N}\ind\ind^\top -\frac{1}{N}\ind\ind^\top =P_m(W_{s+m+1}-\frac{1}{N}\ind\ind^\top )$$
as $P_m$ is doubly stochastic. By the inductive hypothesis, $P_m-\frac{1}{N}\ind\ind^\top =\prod_{\tau =s+1}^{s+m} (W_\tau -\frac{1}{N}\ind\ind^\top )$. Substituting this result in for $P_m$,
\begin{align*}
    P_m(W_{s+m+1}-\frac{1}{N}\ind\ind^\top ) & =\left(\prod_{\tau =s+1}^{s+m} (W_\tau -\frac{1}{N}\ind\ind^\top )+\frac{1}{N}\ind\ind^\top\right) (W_{s+m+1}-\frac{1}{N}\ind\ind^\top ) \\
    & =\frac{1}{N}\ind\ind^\top (W_{s+m+1}-\frac{1}{N}\ind\ind^\top )+\prod_{\tau =s+1}^{s+m+1} (W_\tau -\frac{1}{N}\ind\ind^\top ) \\
    & =\prod_{\tau =s+1}^{s+m+1} (W_\tau -\frac{1}{N}\ind\ind^\top )
\end{align*}
As a result,
$$\norm{E_{s:t-1}}_2 =\norm{\prod_{\tau =s+1}^{t-1} (W_\tau -\frac{1}{N}\ind\ind^\top )}_2\leq\prod_{\tau =s+1}^{t-1} (W_\tau -\frac{1}{N}\ind\ind^\top\leq\rho^{t-s-1}$$
Using the decomposition for $W_t$ we can rewrite $z_{j,k}(t)$.
\begin{align*}
    z_{j,k}(t) & =\frac{1}{N}\ind\ind^\top\sum_{s=1}^{t-1}\Delta\hat{\mu}_{j,k} (s)+\sum_{s=1}^{t-1}E_{s:t}\Delta\hat{\mu}_{j,k} (s) \\
    & =\frac{1}{N}\ind\sum_{i=1}^N\sum_{s=1}^{t-1}\Delta\hat{\mu}_{i,j,k} (s)+\sum_{s=1}^{t-1}E_{s:t}\Delta\hat{\mu}_{j,k} (s) \\
    & =\frac{1}{N}\ind\sum_{i=1}^N\hat{\mu}_{i,j,k} (t-1)+\sum_{s=1}^{t-1}E_{s:t}\Delta\hat{\mu}_{j,k} (s) \\
    & =\ind\bar{\mu}_{j,k} (t-1)+\sum_{s=1}^{t-1}E_{s:t}\Delta\hat{\mu}_{j,k} (s)
\end{align*}
Applying the triangle inequality,
$$\abs{\left[\sum_{s=1}^{t-1}E_{s:t}\Delta\hat{\mu}_{j,k} (s)\right]_i}\leq\sum_{s=1}^{t-1}\norm{E_{s:t}}_2\norm{\Delta\hat{\mu}_{j,k} (s)}_\infty$$
Lemma 1 allows us to conclude,
$$\abs{z_{i,j,k} (t)-\bar{z}_{j,k} (t)} =\abs{\left[\sum_{s=1}^{t-1}E_{s:t}\Delta\hat{\mu}_{j,k} (s)\right]_i}\leq\sum_{s=1}^{t-1}\norm{E_{s:t}}_2\norm{\Delta\hat{\mu}_{j,k} (s)}_\infty\leq\sum_{s=1}^{t-1}\rho^{t-s-1}\norm{\Delta\hat{\mu}_{j,k} (s)}_\infty$$
In Lemma 3, we determine an increment bound for $\norm{\Delta\hat{\mu}_{j,k} (s)}_\infty$. When an update occurs the local average
$$\hat{\mu}_{i,j,k} (s)=\frac{T_{i,j,k}(s-1)\hat{\mu}_{i,j,k} (s-1)+S_{i,j}(s)}{T_{i,j,k}(s)}\text{ where } T_{i,j,k}(s)=T_{i,j,k}(s-1)+1$$
Writing $\Delta\hat{\mu}_{i,j,k} (s)$ in terms of this update
\begin{align*}
    \Delta\hat{\mu}_{i,j,k} (s) & =\hat{\mu}_{j,k} (s)-\hat{\mu}_{j,k} (s-1) \\
    & =\frac{T_{i,j,k}(s-1)\hat{\mu}_{i,j,k} (s-1)+S_{i,j}(s)-T_{i,j,k}(s)\hat{\mu}_{i,j,k} (s-1)}{T_{i,j,k}(s)} \\
    & =\frac{S_{i,j}(s)-\hat{\mu}_{i,j,k} (s-1)}{T_{i,j,k}(s)}
\end{align*}
This implies that
$$\abs{\Delta\hat{\mu}_{i,j,k} (s)} =\frac{\abs{S_{i,j}(s)-\hat{\mu}_{i,j,k} (s-1)}}{T_{i,j,k}(s)}\leq\frac{1}{T_{i,j,k}(s)}$$
Therefore
$$\norm{\Delta\hat{\mu}_{j,k} (s)}_\infty\leq\max_{i\in [N]}\frac{1}{T_{i,j,k}(s)}\text{ and }\abs{z_{i,j,k} (t)-\bar{z}_{j,k} (t)}\leq\sum_{s=1}^{t-1}\rho^{t-s-1}\max_{i\in [N]}\frac{1}{T_{i,j,k}(s)}$$
In the next series of lemmas we consider the sampling error for particular $i,j\in [N]$, $k\in [K]$, and $t\in [T]$. In Lemma 4, we relate the off-diagonal counter $T_{i,j,k}(t)$ to its associated diagonal counter $T_{i,i,k}(t)$. In order for $T_{i,j,k}$ to increase, we must have $a_i^t=k$ and $j\in\mathcal{N}_i (t)$. Conditioning on the history of arms $\mathcal{F}_t =\{ a_i^1,\dots ,a_i^t\}$,
$$T_{i,j,k}(t)\sim\text{Binomial} (T_{i,i,k}(t),p)$$
where $\E [T_{i,j,k}(t)]=pT_{i,i,k}(t)$. Applying Chernoff's bound
$$\Pr [T_{i,j,k}(t)\leq (1-\delta )pT_{i,i,k}(t)]\leq\exp\left( -\frac{\delta^2}{2} pT_{i,i,k}(t)\right)$$
Setting $\delta =\frac{1}{2}$,
$$\Pr [T_{i,j,k}(t)\leq\frac{p}{2} T_{i,i,k}(t)]\leq\exp\left( -\frac{pT_{i,i,k}(t)}{8}\right)$$
Defining a burn-in period $\tau =\frac{32\log (N^2KT)}{p}$, for $T_{i,i,k}(t)\geq\tau$, we have $\frac{pT_{i,i,k}(t)}{8}\geq 4\log (N^2KT)$. Therefore
$$\Pr [T_{i,j,k}(t)\leq\frac{p}{2} T_{i,i,k}(t)]\leq\exp (-4\log (N^2KT))=(N^2KT)^{-4}$$
Taking a union bound over $i,j\in [N]$, $k\in [K]$, and $t\in [T]$,
$$\Pr [\forall i,j,k,t:T_{i,j,k}(t)\leq\frac{p}{2} T_{i,i,k}(t)]\geq 1-N^2KT\cdot (N^2KT)^{-4}=1-(N^2KT)^{-3}$$
Additionally, we can identify a piecewise bound for the off-diagonal counter $T_{i,j,k}(t)$.
$$\frac{1}{\sqrt{T_{i,j,k}(t)}}\leq\begin{cases}
    \sqrt{\frac{2}{p}}\cdot\frac{1}{\sqrt{T_{i,i,k}(t)}} & \text{if } T_{i,i,k}(t)\geq\tau \\
    1 & \text{otherwise}
\end{cases}$$
In Lemma 5, we determine a local concentration bound on the local averages. Suppose that the update times for agent $i$ estimating the value of arm $k$ with respect to agent $j$'s preferences are given by the sequence $\tau_1,\dots ,\tau_m$ where $\tau_m =T_{i,j,k}(t)$. Then,
$$\hat{\mu}_{i,j,k} (t)=\frac{1}{m}\sum_{r=1}^m S_{i,j} (\tau_r )$$
where $S_{i,j} (\tau_r )\in [0,1]$ are iid with $\E [S_{i,j} (\tau_r )]=\mu_{i,j,k}^*$. By Hoeffding's bound
$$\Pr [\abs{\frac{1}{m}\sum_{r=1}^m S_{i,j}(\tau_r )-\mu_{i,j,k}^*}\geq\epsilon ]\leq 2\exp (-2m\epsilon^2)$$
Setting $\epsilon =C\sqrt{\frac{\log (N^2Kt)}{m}}$,
$$2\exp (-2m\epsilon^2 )=2\exp (-2C^2\log (N^2Kt))=2(N^2Kt)^{-2C^2}$$
We can choose a $C$ such that with probability at least $1-(N^2Kt)^{-3}$,
$$\abs{\hat{\mu}_{i,j,k}(t) -\mu_{i,j,k}^*}\leq C\sqrt{\frac{\log (N^2Kt)}{T_{i,j,k}(t)}}$$
In Lemma 6, we derive a bound for the sampling error. Using Lemma 1,
$$\abs{\bar{z}_{j,k} (t)-\bar{\mu}_{j,k}} =\abs{\frac{1}{N}\sum_{i=1}^N (\hat{\mu}_{i,j,k} (t-1)-\mu_{i,j,k}^* )}$$
By the triangle inequality,
$$\abs{\frac{1}{N}\sum_{i=1}^N (\hat{\mu}_{i,j,k} (t-1)-\mu_{i,j,k}^* )}\leq\frac{1}{N}\sum_{i=1}^N\abs{\hat{\mu}_{i,j,k} (t-1)-\mu_{i,j,k}^*}$$
Using Lemma 5, with probability at least $1-(N^2Kt)^{-3}$,
$$\frac{1}{N}\sum_{i=1}^N\abs{\hat{\mu}_{i,j,k} (t-1)-\mu_{i,j,k}^*}\leq\frac{1}{N}\sum_{i=1}^N C\sqrt{\frac{\log (NKt)}{T_{i,j,k}(t-1)}}$$
Applying Lemma 4, with probability at least $1-(N^2Kt)^{-3}$,
$$\frac{1}{N}\sum_{i=1}^N C\sqrt{\frac{\log (NKt)}{T_{i,j,k}(t-1)}}\leq\frac{C\sqrt{\log (NKt)}}{N}\left(\sum_{i=1}^N\sqrt{\frac{2}{p}}\cdot\frac{1}{\sqrt{T_{i,i,k}(t-1)}} +\ind_{T_{i,i,k} (t-1)<\tau}\right)$$
Now that we have developed bounds for consensus and sampling error, it remains to relate these errors to NSW regret. To make this connection, we first show a preliminary fact. Suppose that we have sequences $a_1,\dots ,a_N$ and $b_1,\dots ,b_N$. We also introduce sets $V_1,\dots ,V_N$ such that $V_j=\{ a_1,\dots a_j,b_{j+1},\dots b_N\}$. Using the construction 
$$\prod_{j=1}^N a_j-\prod_{j=1}^N b_j=\prod_{u\in V_N} u-\prod_{v\in V_0} v$$
We can equate the following expression to a certain telescoping sum
$$\prod_{u\in V_N} u-\prod_{v\in V_0} v=\sum_{j=1}^N\left(\prod_{u\in V_j} u-\prod_{v\in V_{j-1}} v\right)$$
Notice the two products within the telescoping sum share the same factors expect for $a_j$ and $b_j$. The shared terms can be factored out resulting in
$$\sum_{j=1}^N\left(\prod_{u\in V_j} u-\prod_{v\in V_{j-1}} v\right) =\sum_{j=1}^N\prod_{l<j} a_l\prod_{l>j} b_l(a_j-b_j)$$
From here, we can rewrite the NSW regret in term of deviations between the global averages and the centered arm means. Let $u_{i,j}^t$ be the utility for the $j$th agent with respect to the global averages and $u_j^*$ be the utility for the $j$th agent with respect to the centered arm means.
\begin{align*}
    \abs{\text{NSW} (p,\bar{\mu} )-\text{NSW} (p,z_i(t) )} & =\abs{\sum_{j=1}^N (u_j^*-u_{i,j}^t)\prod_{l<j} u_l^*\prod_{l>j}u_{i,j}^t} \\
    & \leq\sum_{j=1}^N\abs{u_j^*-u_{i,j}^t}\prod_{l<j}\abs{u_l^*}\prod_{l>j}\abs{u_{i,j}^t} \\
    & \leq\sum_{j=1}^N\abs{u_j^*-u_{i,j}^t} \\
    & \leq\sum_{j=1}^N\sum_{k=1}^K p_k\abs{z_{i,j,k}(t)-\bar{\mu}_{j,k}} \\
\end{align*}
This result allows us to bound the deviation between the NSW under the centered arm means and the NSW under the global average matrix of agent $i$. For any $p\in\Delta_K$,
\begin{align*}
    \abs{\text{NSW} (p,\bar{\mu} )-\text{NSW} (p,z_i(t))} & \leq\sum_{j=1}^N\sum_{k=1}^K p_k\abs{z_{i,j,k}(t)-\bar{\mu}_{j,k}} \\
    & \leq\sum_{j=1}^N\sum_{k=1}^K p_k\abs{z_{i,j,k}(t)-\bar{z}_{j,k} (t)} +\abs{\bar{z}_{j,k} (t)-\bar{\mu}_{j,k}}
\end{align*}
Focusing on the sampling error term, applying Lemma 6
\begin{align*}
    \sum_{j=1}^N\sum_{k=1}^K p_k\abs{\bar{z}_{j,k} (t)-\bar{\mu}_{j,k}} & \leq\sum_{j=1}^N\sum_{k=1}^K p_k\frac{C\sqrt{\log (NKt)}}{N}\left(\sum_{i'=1}^N\sqrt{\frac{2}{p}}\cdot\frac{1}{\sqrt{T_{i',i',k}(t-1)}} +\ind_{T_{i',i',k}(t-1)<\tau}\right) \\
    & =C\sqrt{\log (NKt)}\sum_{i'=1}^N\sum_{k=1}^K p_k\left(\sqrt{\frac{2}{p}}\cdot\frac{1}{\sqrt{T_{i',i',k}(t-1)}} +\ind_{T_{i',i',k}(t-1)<\tau}\right) \\
    & =C\sqrt{\log (NKt)}\left(\sum_{i'=1}^N\sum_{k=1}^K p_k\sqrt{\frac{2}{p}}\cdot\frac{1}{\sqrt{T_{i',i',k}(t-1)}} +\sum_{i'=1}^N\sum_{k=1}^K p_k\ind_{T_{i',i',k}(t-1)<\tau}\right)
\end{align*}
Choosing $i=\argmin_{i^*\in [N]} T_{i',i',k} (t-1)$ and using that indicators variables are at most one, we obtain the simplified bound
$$CN\sqrt{\log (NKt)}\left(\sqrt{\frac{2}{p}}\sum_{k=1}^K p_k\cdot\frac{1}{\sqrt{T_{i,i,k}(t-1)}} +\ind_{T_{i,i,k}(t-1)<\tau}\right)$$
Looking at the consensus error term and applying Lemma 3,
$$\sum_{j=1}^N\sum_{k=1}^K p_k\abs{z_{i,j,k}(t)-\bar{z}_{j,k} (t)}\leq\sum_{j=1}^N\sum_{k=1}^K p_k\sum_{s=1}^{t-1}\rho^{t-s-1}\norm{\Delta\hat{\mu}_{j,k} (s)}_\infty$$
Combining our two results,
$$\abs{\text{NSW} (p,\bar{\mu} )-\text{NSW} (p,z_i(t))}\leq CN\sqrt{\log (NKt)}\sqrt{\frac{2}{p}}\sum_{k=1}^K p_k\cdot\frac{1}{\sqrt{T_{i,i,k}(t-1)}} +\Gamma_{t-1}$$
where 
$$\Gamma_{t-1} =CN\sqrt{\log (NKt)}\sum_{k=1}^K p_k\ind_{T_{i,i,k}(t-1)<\tau} +\sum_{j=1}^N\sum_{k=1}^K p_k\sum_{s=1}^{t-1}\rho^{t-s-1}\norm{\Delta\hat{\mu}_{j,k} (s)}_\infty$$
For simplicity, we can rewrite the sampling error in terms of $T_{i,i,k}(t)$ instead of $T_{i,i,k}(t-1)$. In the worst case, we have $T_{i,i,k}(t)=T_{i,i,k}(t-1)+1$. For $T_{i,i,k}(t)\geq 2$, we can show that
$$\frac{1}{\sqrt{T_{i,i,k}(t-1)}}\leq\frac{\sqrt{2}}{\sqrt{T_{i,i,k}(t)}}$$
We can rewrite the bound as
$$\abs{\text{NSW} (p,\bar{\mu} )-\text{NSW} (p,z_i(t))}\leq\frac{2CN}{\sqrt{p}}\sqrt{\log (NKt)}\sum_{k=1}^K p_k\cdot\frac{1}{\sqrt{T_{i,i,k}(t)}} +\Gamma_{t-1}$$
In Lemma 7 we construct a clean event $C_i^t$.
$$C_i^t=\{ p\in\Delta_K :\abs{\text{NSW} (p,\bar{\mu} )-\text{NSW} (p,z_i(t))}\leq\alpha_t\sum_{k=1}^K p_k\sqrt{\frac{\log (NKt)}{T_{i,i,k}(t)}} +\Gamma_{t-1}\}$$
where $\alpha_t =\frac{CN}{\sqrt{p}}$ collapsing $2C$ into $C$. Applying Lemma 6, $\Pr [C_i^t]\geq 1-O((N^2KT)^{-3})$. The sampling error resembles the explore term in the UCB objective. Now, we relate the bound for $\abs{\text{NSW} (p,\bar{\mu} )-\text{NSW} (p,z_i(t))}$ to a bound for $\abs{\text{NSW} (p^*,\bar{\mu} )-\text{NSW} (p_i^t,\bar{\mu} )}$ the deviation of interest for regret. For the sake of clarity, we set 
$$\text{explore}_1 (p)=\alpha_t\sum_{k=1}^K p_k\sqrt{\frac{\log (NKt)}{T_{i,i,k}(t)}}$$ 
We also break up $\Gamma_{t-1}$ so that 
$$\text{explore}_2 (p)=CN\sqrt{\log (NKt)}\sum_{k=1}^K p_k\ind_{T_{i,i,k}(t-1)<\tau}$$ 
and 
$$\text{cons} (p)=\sum_{j=1}^N\sum_{k=1}^K p_k\sum_{s=1}^{t-1}\rho^{t-s-1}\norm{\Delta\hat{\mu}_{j,k} (s)}_\infty$$ 
We can relate $\text{NSW} (p^*,\hat{\mu} )$ to $\text{NSW} (p^*,z_i(t))$ with the clean event in Lemma 7.
$$\text{NSW} (p^*,\bar{\mu} )\leq\text{NSW} (p^*,z_i(t))+\text{sample}_1 (p^*)+\text{sample}_2 (p^*)+\text{cons} (p^*)$$
Next, we relate $\text{NSW} (p^*,z_i(t))$ to $\text{NSW} (p_i^t,z_i(t))$ using the fact that our algorithm optimizes the objective function.
$$\text{NSW} (p^*,z_i(t))+\text{sample}_1 (p^*)\leq\text{NSW} (p_i^t,z_i(t))+\text{sample}_1 (p_i^t)$$
Finally, we can relate $\text{NSW} (p_i^t,\hat{\mu} )$ to $\text{NSW} (p_i^t,z_i(t))$ with the clean event in Lemma 7.
$$\text{NSW} (p_i^t,z_i(t))\leq\text{NSW} (p_i^t,\bar{\mu} )+\text{sample}_1 (p_i^t)+\text{sample}_2 (p_i^t)+\text{cons} (p_i^t)$$
Chaining inequalities, we can relate $\text{NSW} (p^*,\bar{\mu} )$ to $\text{NSW} (p_i^t,\bar{\mu} )$.
$$\text{NSW} (p^*,\bar{\mu} )\leq\text{NSW} (p_i^t,\bar{\mu} )+\text{sample}_2 (p^*)+\text{cons} (p^*)+2\text{sample}_1 (p_i^t)+\text{sample}_2 (p_i^t)+\text{cons} (p_i^t)$$
Thus, we can reformulate NSW regret in term of the sample and consensus errors.
$$R_T=\sum_{i=1}^N\sum_{t=1}^T\text{NSW} (p^*,\bar{\mu} )-\text{NSW} (p_i^t,\bar{\mu} )\leq\sum_{i=1}^N\sum_{t=1}^T\text{sample}_2 (p^*)+\text{cons} (p^*)+2\text{sample}_1 (p_i^t)+\text{sample}_2 (p_i^t)+\text{cons} (p_i^t)$$
We bound each of the error terms individually starting with the sampling error 1. Before proceeding, in Lemma 8, we will prove a helpful result which bounds with high probability the time an arm counter stays at a specific value. Fix an agent $i$ and arm $k$. We are interested in the count for agent $i$ pulling arm $k$, $T_{i,i,k}(t)$. Let $S_{s:u}=\sum_{t=s}^up_{i,k}^t$ be the sum of the probability that agent $i$ pulls arm $k$ from time $s$ to $u$. Suppose $E_{s:u}$ is the event that arm $k$ is not pulled in the time frame from $s$ to $u$. Conditioning on the history of arm pulls, $\Pr [a_i^t\neq k|\mathcal{F}_{t-1} ]=1-p_{i,k}^t$. Leveraging independence $\Pr [E_{s:u}]=\prod_{t=s}^u (1-p_{i,k}^t)$. Using the identity $1-x\leq e^{-x}$ we can bound the probability
$$\Pr [E_{s:u}]=\prod_{t=s}^u (1-p_{i,k}^t)\leq\exp (-\sum_{t=s}^u p_{i,k}^t)=e^{-S_{s:u}}$$
Let $\tau_l$ be the first time $T_{i,i,k}(t)=l$ and $I_l=[\tau_l ,\tau_{l+1} -1]$, the interval that $T_{i,i,k}(t)=l$. Suppose that $A_l=\sum_{t\in I_l} p_{i,k}^t$ is the sum of the probability that agent $i$ pulls arm $k$ when $T_{i,i,k}(t)=l$. Using the previous result,
$$\Pr [A_l\geq B\log T]\leq e^{-A_l}\leq e^{-B\log T}=T^{-B}$$
Union bounding the $T$ phase for all counters setting $B\geq 3$
$$\Pr [\exists l:A_l\geq B\log T]\leq T\cdot T^{-B}=T^{1-B}$$
$$\Pr [\forall l:A_l\leq B\log T]\geq 1-T^{-2}$$
We will split the bound of sampling error 1 into a warm up regime and an asymptotic regime.
\begin{align*}
    \sum_{i'=1}^N\sum_{t=1}^T\text{sample}_1 (p_{i'}^t) & \leq\frac{CN}{\sqrt{p}}\sqrt{\log (NKT)}\sum_{i'=1}^N\sum_{t=1}^T\sum_{k=1}^K p_{i',k}^t\frac{1}{\sqrt{T_{i,i,k}(t)}} \\
    & =\frac{CN}{\sqrt{p}}\sqrt{\log (NKT)}\sum_{i'=1}^N\sum_{k=1}^K\sum_{t=1}^T p_{i',k}^t\frac{1}{\sqrt{T_{i,i,k}(t)}} \\
    & \leq\frac{CN}{\sqrt{p}}\sqrt{\log (NKT)}\sum_{i'=1}^N\sum_{k=1}^K\left(\sum_{t:\sum p_{i',k}^t< B\tau\log T} p_{i',k}^t\frac{1}{\sqrt{T_{i,i,k}(t)}} +\sum_{t:\sum p_{i',k}^t\geq B\tau\log T} p_{i',k}^t\frac{1}{\sqrt{T_{i,i,k}(t)}}\right) \\
    & \leq\frac{CN}{\sqrt{p}}\sqrt{\log (NKT)}\sum_{i'=1}^N\sum_{k=1}^K\left(\sum_{l=1}^\tau A_l\cdot\frac{1}{l}+\frac{1}{\sqrt{\tau}}\sum_{t=1}^T p_{i',k}^t\right) \\
    & \leq\frac{CN}{\sqrt{p}}\sqrt{\log (NKT)}\sum_{i'=1}^N\sum_{k=1}^K\left(B\tau\log T+\frac{T}{\sqrt{\tau}}\right) \\
    & =\frac{CN}{\sqrt{p}}\sqrt{\log (NKT)}(NKO(T^{3/4}))
\end{align*}
In the last step, we optimize $\tau$ by setting it to $\sqrt{T}$. Note that in the sum over the warm up and asymptotic regime we have arm probability of agent $i'$ times a term involving the counter for agent $i$, the agent with the lowest counter. We reconcile this in the warm up regime by applying the trivial bound $\frac{1}{\sqrt{T_{i,i,k}(t)}}\leq 1$ and we can bound the sum of arm probabilities for agent $i'$ by $B\log T$ using Lemma 8. In the asymptotic regime, note that for all $i'$, $T_{i',i',k}(t)\geq\tau$ by a converse argument that can proven in the same spirit as Lemma 8. Therefore, $\frac{1}{\sqrt{T_{i,i,k}(t)}}\leq\frac{1}{\sqrt{\tau}}$. To bound $\text{sample}_2 (p_i^t)$, we follow a similar structure to the previous bound relying on Lemma 8.
\begin{align*}
    \sum_{i'=1}^N\sum_{t=1}^T\text{sample}_2 (p_{i'}^t) & \leq CN\sqrt{\log (NKt)}\sum_{i'=1}^N\sum_{t=1}^T\sum_{k=1}^K p_{i',k}^t\ind_{T_{i,i,k}(t-1)<\tau} \\
    & \leq CN\sqrt{\log (NKt)}\sum_{i'=1}^N\sum_{k=1}^K\sum_{t=1}^T p_{i',k}^t\ind_{t:\sum p_{i,k}^t<B\tau\log T} \\
    & \leq CN\sqrt{\log (NKt)}\sum_{i'=1}^N\sum_{k=1}^K\sum_{l=1}^\tau A_l \\
    & \leq CN\sqrt{\log (NKt)}\sum_{i'=1}^N\sum_{k=1}^K B\tau\log T \\
    & = CN\sqrt{\log (NKt)} (NKO(\log^2 T))
\end{align*}
To optimize the bound we choose $\tau=\log T$. We apply the same argument to $\text{sample}_2 (p^*)$.
\begin{align*}
    \sum_{i'=1}^N\sum_{t=1}^T\text{sample}_2 (p^*) & \leq CN\sqrt{\log (NKt)}\sum_{i'=1}^N\sum_{t=1}^T\sum_{k=1}^K p_k^*\ind_{T_{i,i,k}(t-1)<\tau} \\
    & \leq CN\sqrt{\log (NKt)}\sum_{i'=1}^N\sum_{t=1}^T\ind_{t:\sum p_{i,k}^t<B\tau\log T} \\
    & \leq CN\sqrt{\log (NKt)}\sum_{i'=1}^N\sum_{k=1}^K\tau \\
    & \leq CN\sqrt{\log (NKt)}(NKO(\log T))
\end{align*}
Finally, we bound $\text{cons} (p_i^t)$ and $\text{cons} (p^*)$. As we observe later, we need not differentiate between $p_i^t$ and $p^*$.
\begin{align*}
    \sum_{i=1}^N\sum_{t=1}^T\text{cons} (p) & \leq\sum_{i=1}^N\sum_{t=1}^T\sum_{j=1}^N\sum_{k=1}^K p_k\sum_{s=1}^{t-1}\rho^{t-s-1}\norm{\Delta\hat{\mu}_{j,k} (s)}_\infty \\
    & =\sum_{i=1}^N\sum_{j=1}^N\sum_{k=1}^K p_k\sum_{s=1}^{T-1}\norm{\Delta\hat{\mu}_{j,k} (s)}_\infty\sum_{t=s+1}^T\rho^{t-s-1} \\
    & \leq\frac{1}{1-\rho}\sum_{i=1}^N\sum_{j=1}^N\sum_{k=1}^K p_k\sum_{s=1}^{T}\norm{\Delta\hat{\mu}_{j,k} (s)}_\infty \\
    & \leq\frac{1}{1-\rho}\sum_{i=1}^N\sum_{j=1}^N\sum_{k=1}^K p_k\sum_{l=1}^N\sum_{s=1}^{T}\abs{\Delta\hat{\mu}_{l,j,k} (s)} \\
\end{align*}
From $s=1$ to $T$, $\abs{\Delta\hat{\mu}_{l,j,k} (s)} >0$ exactly $T_{l,j,k}(T)$ instances. For $1\leq n\leq T_{l,j,k}(t)$, $\abs{\Delta\hat{\mu}_{l,j,k} (s)}\leq\frac{1}{n}$ by Lemma 3.
\begin{align*}
    \frac{1}{1-\rho}\sum_{i=1}^N\sum_{j=1}^N\sum_{k=1}^K p_k\sum_{l=1}^N\sum_{s=1}^{T}\abs{\Delta\hat{\mu}_{l,j,k} (s)} & \leq\frac{1}{1-\rho}\sum_{i=1}^N\sum_{j=1}^N\sum_{k=1}^K p_k\sum_{l=1}^N\sum_{n=1}^{T_{l,j,k}(T)}\frac{1}{n} \\
    & \leq\frac{1}{1-\rho}\sum_{i=1}^N\sum_{j=1}^N\sum_{k=1}^K p_k\sum_{l=1}^N(\log T+1) \\
    & =\frac{1}{1-\rho}\sum_{i=1}^N\sum_{j=1}^N\sum_{k=1}^K p_k(N(\log T+1)) \\
    & =\frac{N^3}{1-\rho} (\log T+1)
\end{align*}
Combining all error terms, we observe that the $\text{sample}_1 (p_{i'}^t)$ term dominates regret. Thus, our final bound on NSW regret is
$$R_T\leq O(\frac{CN}{\sqrt{p}}\sqrt{\log (NKT)}(NKO(T^{3/4})))$$

\subsection{Experimental Reproducibility Details}\label{reprod}

Experiments were run on a Dell XPS 13 (9315) with a 12th Gen Intel Core i7-1250U CPU (1.10 GHz) and 32 GB RAM. All code was executed in Jupyter Notebook environments, with runtimes of approximately 2 hours and 1.5 hours for the two experimental settings. To facilitate reproducibility, the source code is available in an anonymized GitHub repository: \url{https://anonymous.4open.science/r/Pareto-NSW-MO-MA-MAB-Experiments-04FC/}.

\end{document}